\documentclass[journal]{IEEEtran}
\ifCLASSINFOpdf
\else
\fi

\usepackage{subfigure}
\usepackage{epsfig}
\usepackage{graphicx}
\usepackage{amsmath}
\usepackage{amssymb}
\usepackage{algorithm, algorithmic}
\graphicspath{{figure/}}   
\usepackage{multirow}

\usepackage{hyperref}
\usepackage{booktabs}
\usepackage{color}
\usepackage{bbding}
\usepackage{tabularx}
\begin{document}
%
\title{Deep Intrinsic Decomposition with Adversarial Learning for Hyperspectral Image Classification}
%
%
%

\author{Zhiqiang~Gong,		
			Xian~Zhou,
			Wen~Yao
\thanks{Manuscript received XX, 2023; revised XX, 2023. This work was supported by the Natural Science Foundation of China under Grant 62001502. 
}
\thanks{ Z. Gong and W. Yao are with the Defense Innovation Institute,
Chinese Academy of Military Sciences, Beijing 100071, China. e-mail: (gongzhiqiang13@nudt.edu.cn, wendy0782@126.com).}
\thanks{X. Zhou is with  the Information Research Center of Military Science, Chinese Academy of Military Sciences, Beijing 100000, China (e-mail: zhouxian@sjtu.edu.cn).}

}

%
%

\markboth{IEEE LATEX,~Vol X, 2023}%
{Shell \MakeLowercase{\textit{et al.}}: Bare Demo of IEEEtran.cls for IEEE Journals}
%



\maketitle

\begin{abstract}
Convolutional neural networks (CNNs) have been demonstrated their powerful ability to extract discriminative features for hyperspectral image classification. However, general deep learning methods for CNNs ignore the influence of complex environmental factor which enlarges the intra-class variance and decreases the inter-class variance. This multiplies the difficulty to extract discriminative features.
To overcome this problem, this work develops a novel deep intrinsic decomposition with adversarial learning, namely AdverDecom, for hyperspectral image classification to mitigate the negative impact of environmental factors on classification performance.
First, we develop a generative network for hyperspectral image (HyperNet) to extract the environmental-related feature and category-related feature from the image. Then, a discriminative network is constructed to distinguish different environmental categories. Finally, a environmental and category joint learning loss is developed for adversarial learning to make the deep model learn discriminative features.
Experiments are conducted over three commonly used real-world datasets and the comparison results show the superiority of the proposed method.
The implementation of the proposed method and other compared methods could be accessed at \url{https://github.com/shendu-sw/Adversarial_Learning_Intrinsic_Decomposition} for the sake of reproducibility.
\end{abstract}


%
\IEEEpeerreviewmaketitle

\section{Introduction}

Hyperspectral images, which contain a multitude of spectral bands including the visible and non-visible parts of the electromagnetic spectrum \cite{gong3}, can provide an extensive and detailed view of the Earth's surface and 
play a crucial role in various fields, including agriculture, geology, ecology, and disaster management \cite{b19}.
The plentiful spectral and spatial information of hyperspectral data allows for precise discrimination and characterization of materials, terrain, and environmental features, facilitating applications such as land cover mapping \cite{b20}, mineral identification \cite{b21}, vegetation health assessment \cite{b22}, and pollution monitoring \cite{b23}. 
However, great spectral similarity occurs between different objects which makes difficulty to discriminate different objects. Another challenge arises from the complexity of handling a vast amount of spectral information across numerous narrow bands. The high dimensionality of the data poses difficulties in effective feature selection, model training, and computational demands. Additionally, atmospheric effects, mixed pixels, and the need for extensive, accurately labeled training data make hyperspectral classification a formidable task.
Therefore, there exists huge demand to explore effective methods to extract discriminative features from the hyperspectral image.

\begin{enumerate}
\item A robust and effective feature extraction backbone network to understand and represent the complex spectral-spatial correlation of hyperspectral image is required. Generally, a well-designed network would have great potential to capture relevant patterns and characteristics from the training samples.
\item A proper learning strategy is imperative to truly harness the discriminative information of the hyperspectral image. Especially, by considering the unique characteristics of the hyperspectral image, the model's representational ability could be enhanced to extract valuable latent information from the complex data.
\end{enumerate}
Following these two fundamental considerations, there have been increasing efforts to explore impressive methods for hyperspectral image classification.

Faced with the first problem, conventional methods design hand-crafted spectral features to represent the hyperspectral image. These well-established techniques usually includes spectral feature extraction (e.g. principal component analysis (PCA) \cite{b24}, linear discriminant analysis (LDA) \cite{b25}), statistical classifiers \cite{b26}, and dimensionality reduction (e.g. non-negative matrix factorization (NMF) \cite{b27}, t-distributed stochastic neighbor embedding (t-SNE) \cite{b28}, which cannot be adaptive for the complex latent correlation within the image. 
To pursue more representative methods, 
much efforts have been paid on machine learning algorithms, including support vector machines (SVM) \cite{b29}, decision trees \cite{b30}, k-Nearest Neighbors (k-NN) \cite{b31}, and random forests \cite{b32}, to optimize feature extraction and classification processes.  These methods are generally ``shallow'' methods with only one or two layers, which limit their ability to capture the intricate patterns and spectral information embedded in hyperspectral data.

Recently, deep learning with multiple hidden layers have gained prominence in hyperspectral image classification \cite{b18}. 
They can automatically extract hierarchical features from the data and capture complex relationships, which can further enhance classification accuracy.
Generally, based on different architectural paradigms, these deep learning methods can be broadly several classes, such as recurrent neural networks (RNNs), graph convolutional networks (GCNs), CNNs, Transformers, and others. RNNs are good at captuing temporal and spectral dependencies within the hyperspectral data. As a representative, Hang et al. designed a cascaded RNN for HS image classification by taking advantage of RNNs that can model the sequentiality to represent the relations of neighboring spectral bands effectively \cite{b9}. GCNs can effectively capture and propagate information across this spectral graph  \cite{gong5}, allowing for the modeling of complex relationships and contextual dependencies within hyperspectral data. MiniGCN \cite{b5}, which provides a feasible solution
for addressing the issue of large graphs in GCNs, is a representative of this class of methods.
Transformers excel at capturing long-range dependencies in the data, which is especially useful when hyperspectral information is distributed across a wide spectral range. 
ViT \cite{b1}, Transformer in Transformer (TNT) \cite{b2}, SpectralFormer \cite{b3}, are typical transformers which can be applied for hyperspectral image classification.
CNNs, as the most used deep architectures for hyperspectral image classification, can capture local spatial relationships while efficiently process the spectral information. The representative CNNs, such as Noise CNN \cite{gong4}, HybridSN \cite{b6}, PResNet \cite{b7}, and 3-D CNN \cite{b8}, can make full use of both the spatial and spectral information, and present comparable or even better performance than other paradigms.

While these architectures exhibit promising potential for hyperspectral images, they tend to overlook the intrinsic properties of hyperspectral images, thereby limiting their classification performance.
Through analysis of intrinsic structure of hyperspectral image, this work will mainly propose a deep intrinsic decomposition framework for hyperspectral image classification. The framework constains the generative network (HyperNet) and discriminative network to extract the environmental-related features and category-related features, which can mitigate the influence of environmental factors and better discriminate different objects.

In order to deal with the second one, this work develops a novel adversarial learning method for deep intrinsic decomposition utilize the intrinsic physical property. Prior works mainly focus on design the specific training losses to learn a better model.  The common training loss quantifies the disparity between the predicted outputs of the model and the actual ground truth labels during the training process, such as the generally used softmax loss \cite{b35}. Some other works also construct training loss functions for hyperspectral remote sensing images by incorporating inter-sample relationships \cite{b33, b34}. This approach harnesses the spectral similarities and differences between samples in the dataset to improve the performance of deep models. Furthermore, more advanced avenue of research explores the incorporation of the physical properties inherent to categories within hyperspectral data for the construction of training loss functions, such as Statistical loss \cite{gong2},  DMEM loss \cite{gong1}. By considering the unique spectral characteristics and physical attributes of materials or objects of the same category, it becomes possible to design loss functions that promote a deeper understanding and better exploitation of these intrinsic properties. 
While these training loss functions for developing hyperspectral remote sensing image classification models have become capable of harnessing intra-class structural information, they still disregard the influence of environmental factors on hyperspectral imaging.

When dealing with hyperspectral images, it is essential to acknowledge the significant impact that environmental factors have on classification performance. The intricate interplay of these factors can introduce variations in spectral signatures, potentially leading to misclassification or reduced accuracy in the analysis of hyperspectral data.
Researchers try to isolate the unique spectral characteristics or intrinsic properties of the materials or objects within the image through hyperspectral intrinsic decomposition \cite{b13}. However, prior works on hyperspectral intrinsic decomposition pjredominantly relied on general spectral analysis techniques \cite{b14, b15, b16}. The classification performance is limited due to limitations in their model's ability to express complex spectral information effectively.

Motivated by \cite{b10}, this work endeavors to implement deep intrinsic decomposition by leveraging a dedicated adversarial learning method.  The intention is to harness the power of deep neural networks to capture the intricate interplay between spectral and spatial information in hyperspectral data. By incorporating adversarial learning, which involves the training of a generator and discriminator network, the model can learn to disentangle intrinsic components more effectively.

Considering the merits of both the hyperspectral deep intrinsic decomposition and adversarial learning, this work develops a new deep intrinsic decomposition with adversarial learning for hyperspectral image classification.
First, we design a adversarial network which contains the hypernet and discriminative network to extract the environmental-related feature and category-related feature.
Then, a environmental and category joint learning loss is developed for adversarial learning of the model.
Finally, we have successfully implemented deep intrinsic decomposition through our specific adversarial learning framework. 
To be concluded, this paper makes the following contributions.
\begin{itemize}
\item We revisit the intrinsic property of hyperspectral image and propose a new adversarial network  comprising a hypernet and a discriminative network that jointly extract environmental-related and category-related features from hyperspectral data. This innovation enables a more comprehensive understanding of complex scenes.

\item We develop a new adversarial learning based on the environmental and category joint learning loss to make the model learn discriminative environmental-related features and category-related features. This loss function encourages the effective disentanglement of intrinsic components, thereby improving the model's performance in hyperspectral decomposition tasks.

\item We qualitatively and quantitatively evaluate the classification performance of the proposed AdverDecom on three representative hyperspectral image datasets, i.e., Pavia University data, Indian Pines data, and Houston2013 data. Comparisons with other state-of-the-art methods show that the proposed method can have a significant superiority (with an increase of at least 3\% OA).

\end{itemize}

The remainder of this paper is organized as follows. Section \ref{sec:proposed} details the proposed AdverDecom, including hyperspectral intrinsic decomposition, adversarial network and adversarial learning for deep intrinsic decomposition, and implementation details, for hyperspectral image classification. Extensive experiments are conducted over three real-world datasets for quantitative and qualitative evaluation of the proposed method in Section \ref{sec:experiments}.  Section \ref{sec:conclusions} concludes the work with a brief outlook on future directions.

\section{Proposed Method}\label{sec:proposed}

Given a specific hyperspectral image, the goal of classification task is to assign a unique land-cover label to each pixel of the image.
Denote $X=\{{\bf x}_1, {\bf x}_2, \cdots, {\bf x}_N\}$ as the set of training samples of a given hyperspectral image, where $N$ is the number of training samples, and $y_i$ is the corresponding label of ${\bf x}_i$. $y_i \in \Gamma=\{1,2,\cdots, \Lambda\}$ where $\Lambda$ represents the class number of the  image.

\subsection{Hyperspectral Intrinsic Decomposition}\label{subsec:decom}

The intrinsic information coupling model is designed to model the mutual coupling process of intensity and color of light during the imaging process. This model aims to elucidate the intricate interplay between light intensity and color, providing valuable insights into the underlying dynamics of image formation.

As for a natural red-green-blue (RGB) image, 
the intrinsic image decomposition can be described as \cite{b11, b12}
\begin{equation}
I=R\circ S
\end{equation}
where $I$ denotes the original image, $R$ and $S$ represents the reflectance component and the shading component, respectively. $\circ$ stands for the elementwise multiplication operator.

In contrast to RGB images, hyperspectral images are typically acquired using passive imaging sensors that primarily capture energy reflected from solar radiation. Due to variations in sensitivity to scene radiance changes across different spectral bands, the pixel values in different bands undergo non-proportional changes with scene radiance variations. 
Therefore, the shading component of hyperspectral images affects each wavelength differently. Considering the 
varying effects, the hyperspectral intrinsic decomposition model can be formulated as \cite{b17}
\begin{equation}\label{eq:02}
I(\lambda)=R(\lambda) \circ S(\lambda),
\end{equation}
where $\lambda$ denotes the wavelength, $R(\lambda)$ and $S(\lambda)$ represents the reflectance component and the shading component.
$R(\lambda)$ determines the spectral reflectance signature which is unique spectral response of each pixel in the image. $S(\lambda)$ describes the influence of environmental factors on the hyperspectral image. 
Based on the property, we define $R(\lambda)$ and $S(\lambda)$ as the category-related feature and environmental-related feature, respectively.

For the task at hand, 
the objective of hypersepctral intrinsic image decomposition is to decrease the influence of complex environmental factors, extract and represent the intrinsic spectral and spatial information of hyperspectral images accurately, so as to improve the performance of hypersepctral image classification.
Following we will introduce proposed deep intrinsic decomposition method based on the assumption model.

\subsection{Adversarial Network for Deep Intrinsic Decomposition}

\begin{figure*}[t]
\centering
\includegraphics[width=0.99\linewidth]{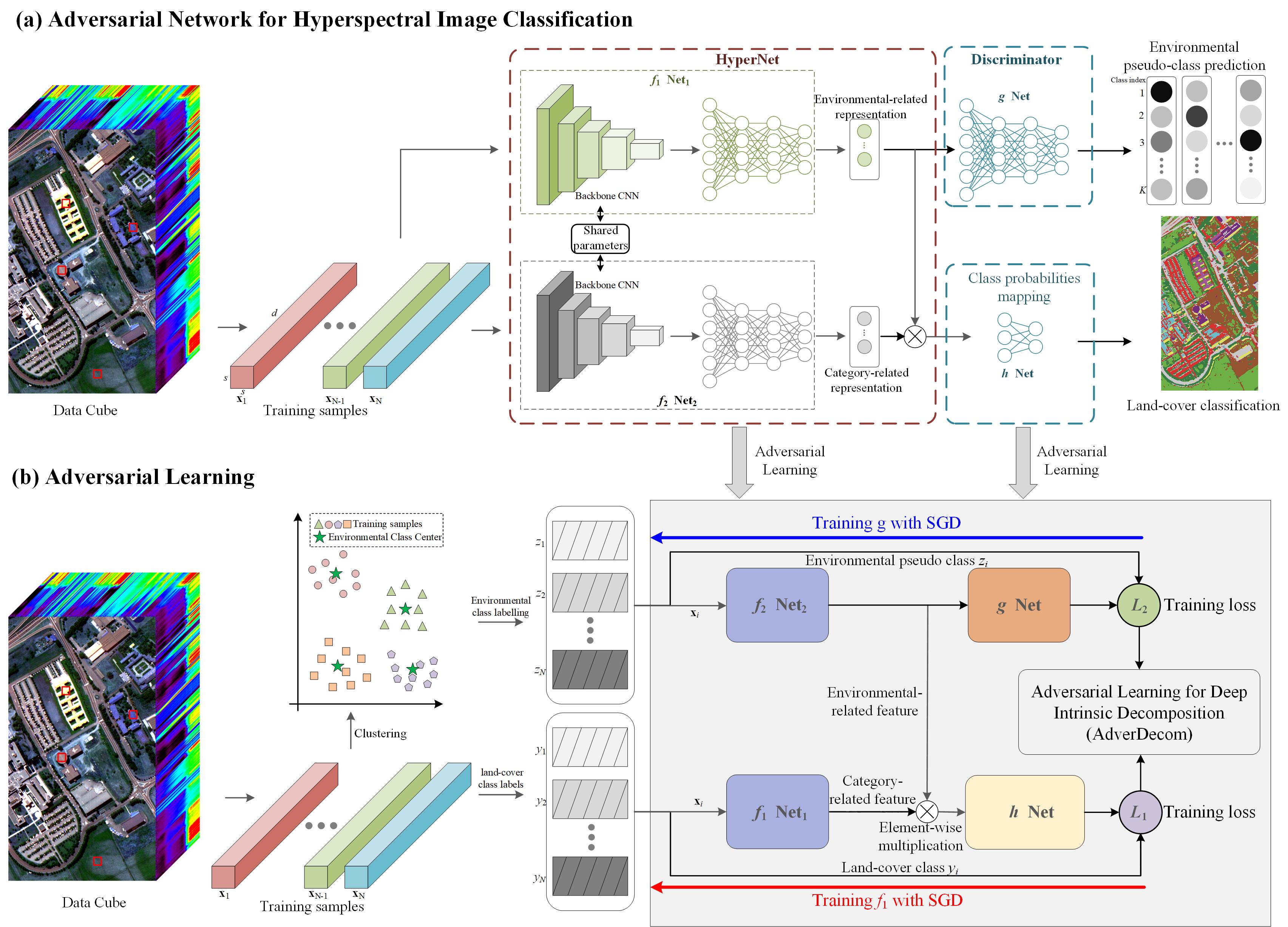}
   \caption{Flowchart of Deep Intrinsic Decomposition with Adversarial Learning (AdverDecom) method for hyperspectral image classification. (a) The adversarial network for hyperspectral image classification. Our network decompose the samples into the environmental-related representation and category-related representation to decrease the influence of complex environmental factors and emphasize the most distinctive and informative spectral signatures  in the data for better classification performance. (b) The illustration of our Adversarial Learning for Deep Intrinsic Decomposition. We construct the environmental class labels under clustering and applied Algorithm \ref{algorithm} to train the proposed adversarial network.}
\label{fig:flowchart}
\end{figure*}

As shown in Fig. \ref{fig:flowchart}, this work constructs a novel adversarial network to realize the deep intrinsic decomposition for hyperspectral image.
The adversarial network consists of the HyperNet and the discriminative network.

\subsubsection{HyperNet}
The aim of HyperNet is to decompose the learned feature into the environmental-related and category-related part.
Under the assumption in subsection \ref{subsec:decom}, the original image can be divided into the category-related feature and environmental-related feature.

Given a sample ${\bf x}_i$ in the image. Define $f_1(\cdot)$ as the function to extract the category-related feature and $f_2(\cdot)$ as the function to extract the environmental-related feature. Then, based on Eq. \ref{eq:02}, the problem can be formulated as
\begin{equation}
f({\bf x}_i, \lambda)=f_1({\bf x}_i,\lambda)\circ f_2({\bf x}_i,\lambda)
\end{equation}
where $f(\cdot)$ denotes the overall feature learned from ${\bf x}_i$. $f_1(\cdot)$ and $f_2(\cdot)$ are fundamentally about learning mapping relationships and extracting features from the image. Deep neural networks are widely recognized for their exceptional nonlinear fitting capabilities, making them a prime choice for implementing functions $f_1(\cdot)$ and $f_2(\cdot)$ in this study.
Deep learning enables the parallel processing of different spectral bands in hyperspectral remote sensing imagery, allowing deep models to model all bands simultaneously. 
Moreover, leveraging deep neural networks allows us to harness the complex, hierarchical representations within hyperspectral data, enabling us to capture intricate patterns and relationships. 

Fig. \ref{fig:flowchart} introduces the framework of the developed HyperNet. $Net_1$ and $Net_2$ are used for the learning of $f_1$ and $f_2$, respectively.
The first halves of $Net_1$ and $Net_2$ consist of a common CNN backbone network model with shared parameters, allowing them to collectively extract and learn essential hierarchical features from the image. 
This shared architecture ensures that both networks benefit from a shared understanding of low-level spectral-spatial features present in the data. In the latter halves of these networks, distinct MLP models are employed, which specialize in different objectives. This design enables the networks to leverage the same foundational feature representations while tailoring their respective output layers to extract the environmental-related and the category-related features.

\subsubsection{Discriminative Network}\label{subsubsec:discriminative}
The discriminative network takes the environmental-related features as input and learns to disciminate the environmental pseudo class out of $K$ pre-defined environmental category.
This work uses a specific multi-layer perception as the discriminate network. Denote $g(\cdot)$ as the mapping function of the discriminate network. Then, the extracted features can be formulated as   
$g(f_2({\bf x}_i))$,
where $g(\cdot): \mathbb{R}^{N_1}\rightarrow \mathbb{R}^{N_2}$ is the representation function,  $N_1$ represents the dimension of environmental-related and category-related features from the image, and $N_2$ stands for the dimension of extracted features by the discriminative network.

\subsection{Adversarial Learning for Deep Intrinsic Decomposition}
In this subsection, we will present the methodology and details of learning the representation function $f(\cdot)$ of the image. 
In particular, we will first introduct the goal of deep intrinsic decomposition, motivate the proposed AdverDecom by general adversarial learning, and finally discuss key algorithmic details.

\subsubsection{Deep Intrinsic Decomposition Goal}
Given sample ${\bf x}_i$ from the hyperspectral image, the goal of Deep Intrinsic Decomposition is to learn a representation $f(\cdot)$, such that for any environmental factors, there exists a latent mapping function $f_2$ which allows $f_1({\bf x}_i)\circ f_2({\bf x}_i)$ sufficiently distinctive to distinguish different land-cover classes. 
Formally, an optimal representation, $f_1$, solves the following optimization problem:
\begin{equation}\label{eq:01}
\min\limits_{f_1, f_2}\sum_{i=1}^{N}C_1(h(f_1({\bf x}_i)\circ f_2({\bf x}_i)), y_i)
\end{equation}
where $f_1, f_2: \mathbb{R}^{s\times s\times d} \rightarrow \mathbb{R}^{N_1}$ is the representation function, $h(\cdot): \mathbb{R}^{N_1}\rightarrow \mathbb{R}^{\Lambda}$ is the mapping from the features to classification probabilities and $C_1$ denotes a classification loss function. $s$ represents the spatial size used for better classifition performance and $d$ stands for the channels of the hyerspectral image.
As showed in prior subsection, we use specific deep neural networks (DNNs) to represent $f_1$ and $f_2$, respectively. 

\subsubsection{Adversarial Learning}

One challenge of the optimization in Eq. \ref{eq:01} is the intraspecies spectral variability caused by the environmental factors. These fluctuations can lead to great differences in the spectral signatures of objects or materials belonging to the same category, and multiplies the difficulty to learn $f_1$ and $f_2$.
Due to the good representational ability, the DNN may memorize the distributions of samples from a specific class under different environmental factors. 
The optimization in Eq. \ref{eq:01} may lead to over-fitting and may not properly find an environmentally invariant representation.

To address the aforementioned issue, this work employs adversarial learning framework to separately acquire environmental-related and category-related features. First, based on the training samples, we develop to construct environmental pseudo classes unsupervisedly. Then, based on the environmental pseudo classes, we construct the adversarial optimization problem.

{\bf Construction of Environmental Pseudo Classes}
Given the training samples of the hyperspectral image, we group them into a number of environmental pseudo classes unsupervisedly. 

In hyperspectral imaging, there exists the phenomenon that distinct objects or materials exhibit similar spectral signatures in the hyperspectral data. This intriguing occurrence generally arise due to the presence of the effects of environmental factors. Therefore, directly conducting clustering analysis on hyperspectral pixels can yield valuable insights into the influence of environmental variables on classification. This approach involves grouping pixels with similar spectral properties, potentially revealing patterns of spectral variability driven by environmental factors.

Denote $K$ as the number of predefined environmental pseudo classes. Denote $P_k(k=1,2,\cdots,K)$ as the centers of different environmental factors.
Iteratively, we calculates the centers of the $K$ groups, optimizing the error as follows:
\begin{equation}\label{eq:s1}
\min\limits_{P_1,P_2, \cdots, P_K}\sum_{i=1}^{N}\sum_{k=1}^{K}I(k=\arg\min\limits_{1,2,\cdots,K} \|{\bf x}_i-P_k\|^2)\|{\bf x}_i-P_k\|^2
\end{equation}
where $I(condition)$ denotes the indicative function where $I(\cdot)=1$ if condition is true $I(\cdot)=0$ otherwise.

Given a training sample ${\bf x}_i$ in the hyperspectral image, denote $z_i$ as the corresponding environmental pseudo class of ${\bf x}_i$, then the environmental pseudo class $z_i$ can be obtained by calculating
\begin{equation}\label{eq:s2}
z_i=\arg\min\limits_{1,2,\cdots,K} \|{\bf x}_i-P_k\|^2. 
\end{equation}
For convenience, following we will use ${\bf x}_i^{(z_i)}$ to represent the sample ${\bf x}_i$ with environment index $z_i$.

{\bf Adversarial Optimization}
To solve the intraspecies spectral variability problem, we propose the following adversarial optimization framework:
\begin{equation}\label{eq:maxmin}
\max\limits_{g}\min\limits_{f_1}\sum_{i=1}^{N}(C_1(h(f_1({\bf x}_i)\circ f_2({\bf x}_i)), y_i)-\alpha\cdot C_2(g(f_2({\bf x}_i^{(z_i)})), z_i))
\end{equation}
where $g$ represents the DNN as the discriminator in subsection \ref{subsubsec:discriminative} to predict the environment index out of $K$ environmental pseudo classes.
$C_1(\cdot)$ and $C_2(\cdot)$ represent the classification loss functions (e.g., the cross entropy loss), respectively, $\alpha\geq 0$ is a hyperparameter to control the degree of regularization. 
Intuitively, $g$, and $f_1$ play a zero-sum max-min game: the goal of $g$  is to predict the environmental index $z_i$ directly from $f_1({\bf x}_i)$ (achieved by the outer $\min$; the goal of $f_1$ is to approximate the label $y_i$ while making the job of $g$ harder (achieved by the inner $\max$). In other words, $g$ is a learned regularizer to remove the environmental information contained in $f_1$. 

In our experiments, the output of $h$ is a $\Lambda$-dimensional vector for the class probabilities of $\Lambda$ land-cover classes, and we use the cross entropy loss for $C_1(\cdot)$, which is given as
\begin{equation}
C_1(h(f_1({\bf x}_i)\circ f_2({\bf x}_i)), y_i)=-\sum\limits_{j=1}^{\Lambda} \delta_{jy_i}\log(h(f_1({\bf x}_i)\circ f_2({\bf x}_i))^Te_j)
\end{equation}
where $\delta_{jy_i}=1$ if $j=y_i$ and $\delta_{jy_i}=0$ otherwise and $e_j\in \mathbb{R}^\Lambda$ stands for the standard basis vector.
Similarly, $g$ represents a $K$-dimensional vector for the class probabilities of $K$ land-cover classes, and we also use the cross entropy loss for $C_2(\cdot)$, which is given as
\begin{equation}
C_2(g(f_2({\bf x}_i^{(z_i)})), z_i)=-\sum\limits_{j=1}^{K}\delta_{jz_i}\log(g(f_2({\bf x}_i^{(z_i)}))^Te_j)
\end{equation}
where $\delta_{jz_i}=1$ if $j=z_i$ and $\delta_{jz_i}=0$ otherwise and $e_j\in \mathbb{R}^K$ stands for the standard basis vector.

\subsection{Implementation of Deep Intrinsic Decomposition with Adversarial Learning}

Finally, we solve the optimization problem in \ref{eq:maxmin} by the proposed AdverDecom (described in Algorithm \ref{algorithm} and Fig. \ref{fig:flowchart}). As shows in the algorithm, the AdverDecom contains three steps: (1) Construct the environmental pseudo classes of different samples (Line 4); (2)  update the category-related representation based on the training batch $B$ (Line 5); (3) update the discriminator $g$ on the training batch $B$ (Line 6). Under iteratively learning from (2) to (3), we can obtain the final environmental-invariant features.

\begin{algorithm}[t]
\renewcommand{\algorithmicrequire}{\textbf{Input:}}
\renewcommand{\algorithmicensure}{\textbf{Output:}}
\caption{Deep Intrinsic Decomposition with Adversarial Learning (AdverDecom)}
\begin{algorithmic}[1]
\REQUIRE (${\bf x}_i$, $y_i) (i=1,2,\cdots, N)$, $\alpha$, $K$
\ENSURE Deep neural networks $f_1$, $f_2$, $g$, $h$.
\REPEAT
\STATE Randomly sample training batch B.
\STATE Initialize the deep neural networks $f_1$, $f_2$, $g$, $h$.
\STATE Compute the environmental pseudo classes $z_i$ of different samples ${\bf x}_i$ using Eqs. \ref{eq:s1} and \ref{eq:s2}.
\STATE Train $f_1$ using stochastic gradient descent (SGD) with training loss $L_1$
\begin{equation}
L_1=\sum\limits_{{\bf x}_i^{(z_i)}\in B}(C_1(h(f_1({\bf x}_i)\circ f_2({\bf x}_i)), y_i)-\alpha\cdot C_2(g(f_2({\bf x}_i^{(z_i)})), z_i))
\end{equation}
\STATE Train $g$ using stochastic gradient descent (SGD) with training loss $L_2$
\begin{equation}
L_2=\sum\limits_{{\bf x}_i^{(z_i)}\in B} C_2(g(f_2({\bf x}_i^{(z_i)})), z_i))
\end{equation}
\UNTIL{Convergence}
\end{algorithmic}
\label{algorithm}
\end{algorithm}

\section{Experimental Results}\label{sec:experiments}


\subsection{Experimental Datasets}

The classification performance of the proposed AdverDecomCNN is evaluated on three datasets, i.e., the Pavia University dataset \cite{data}, the Indian Pines dataset \cite{data}, and the Houston2013 dataset \cite{houston2013}.

{\bf Pavia University (PU) data} was obtained by the reflective optics system imaging spectrometer (ROSIS-3) over the city of Pavia, Italy with a spatial resolution of $1.3m\times 1.3m$. It consists of $610\times 340$ pixels and each pixel possesses 115 bands with a spectral coverage ranging from 0.43 to 0.86 $\mu$m. 12 spectral bands are abandoned due to the water absorption and noise, and the remaining 103 channels are used. A total of 43923 labeled sampels divided into nine classes have been chosen for experiments (seen table \ref{table:pavia} for details). The number of training and testing samples per class are also listed in the table.

{\bf Indian Pines (IP) data} was gathered by the Airborne Visible/Infrared Imaging Spectrometer (AVIRIS) sensor over the Indian PInes test set in Northwestern Indiana at a ground sampling distance (GSD) of 20m. It consists of $145\times 145$ pixels with spectral bands  ranging from 0.4 to 2.5 $\mu$m. 24 bands covering the region of water absorption are removed and the remaining 200 spectral bands are used. 16 land cover classes with a total of 10366 labeled samples are selected for experiments. Table \ref{table:indian} shows the detailed training and testining samples in the experiments.

{\bf Houston 2013 (HS) data } was collected by the National Center for Airborne Laser Mapping (NCALM) over the University of Houston campus and the neighboring urban area  throuth ITRES CASI 1500 sensor at the spatial resolution of 2.5m. The cube consists of $349\times 1905$ pixels with 144 spectral bands ranging from 380 nm to 1050 nm. 15 land cover classes with a total of 15029 labeled samples are selected for experiments. Table \ref{table:houston2013} presents the details of the training and testing samples of the dataset for experiments. 


\begin{table}[t]
\begin{center}
\caption{Number of training and testing samples in Pavia University data.}
\label{table:pavia}
\begin{tabular}{ c | c c c c }
\toprule[1pt]
{ Class}     &  { Class Name} & Color &  { Training}&  { Testing} \\
\hline\hline
C1   &  Asphalt                  &\begin{minipage}[b]{0.08\columnwidth}
		\raisebox{-.45\height}{\includegraphics[width=\linewidth]{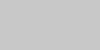}}
	\end{minipage}  & 548  & 6304  \\
C2   &  Meadows                  & \begin{minipage}[b]{0.08\columnwidth}
		\raisebox{-.45\height}{\includegraphics[width=\linewidth]{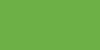}}
	\end{minipage} & 540  & 18146 \\
C3   &  Gravel                   & \begin{minipage}[b]{0.08\columnwidth}
		\raisebox{-.45\height}{\includegraphics[width=\linewidth]{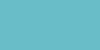}}
	\end{minipage} & 392  & 1815   \\
C4   &  Trees                    & \begin{minipage}[b]{0.08\columnwidth}
		\raisebox{-.45\height}{\includegraphics[width=\linewidth]{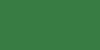}}
	\end{minipage} & 524  & 2912   \\
C5   &  Metal sheet              & \begin{minipage}[b]{0.08\columnwidth}
		\raisebox{-.45\height}{\includegraphics[width=\linewidth]{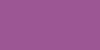}}
	\end{minipage} & 265  & 1113  \\
C6   &  Bare soil                & \begin{minipage}[b]{0.08\columnwidth}
		\raisebox{-.45\height}{\includegraphics[width=\linewidth]{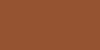}}
	\end{minipage} & 532  & 4572  \\
C7   &  Bitumen                  & \begin{minipage}[b]{0.08\columnwidth}
		\raisebox{-.45\height}{\includegraphics[width=\linewidth]{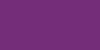}}
	\end{minipage} & 375  & 981   \\
C8   &  Brick                    & \begin{minipage}[b]{0.08\columnwidth}
		\raisebox{-.45\height}{\includegraphics[width=\linewidth]{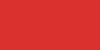}}
	\end{minipage} & 514  & 3364   \\
C9   &  Shadow                   & \begin{minipage}[b]{0.08\columnwidth}
		\raisebox{-.45\height}{\includegraphics[width=\linewidth]{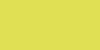}}
	\end{minipage} & 231  & 795  \\

 \hline\hline
Total     &         &  & 3921 & 40002  \\
\bottomrule[1pt]
\end{tabular}
\end{center}
\end{table}

\begin{table}[t]
\begin{center}
\caption{Number of training and testing samples in Indian Pines data.}
\label{table:indian}
\begin{tabular}{ c | c c c c }
\toprule[1pt]
{Class}     &  {Class Name} & Color &  {Training}&  {Testing}  \\
\hline\hline
C1   &  Corn-notill                    &\begin{minipage}[b]{0.08\columnwidth}
		\raisebox{-.45\height}{\includegraphics[width=\linewidth]{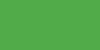}}
	\end{minipage} & 50  & 1384   \\
C2   &  Corn-mintill                   &\begin{minipage}[b]{0.08\columnwidth}
		\raisebox{-.45\height}{\includegraphics[width=\linewidth]{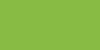}}
	\end{minipage} & 50  & 784  \\
C3   &  Corn                           &\begin{minipage}[b]{0.08\columnwidth}
		\raisebox{-.45\height}{\includegraphics[width=\linewidth]{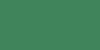}}
	\end{minipage} & 50  & 184   \\
C4   &  Grass-pasture                  &\begin{minipage}[b]{0.08\columnwidth}
		\raisebox{-.45\height}{\includegraphics[width=\linewidth]{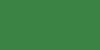}}
	\end{minipage} & 50  & 447   \\
C5   &  Grass-trees                    &\begin{minipage}[b]{0.08\columnwidth}
		\raisebox{-.45\height}{\includegraphics[width=\linewidth]{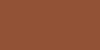}}
	\end{minipage} & 50  & 697  \\
C6   &  Hay-windrowed                  &\begin{minipage}[b]{0.08\columnwidth}
		\raisebox{-.45\height}{\includegraphics[width=\linewidth]{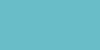}}
	\end{minipage} & 50  & 439   \\
C7   &  Soybean-notill                 &\begin{minipage}[b]{0.08\columnwidth}
		\raisebox{-.45\height}{\includegraphics[width=\linewidth]{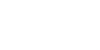}}
	\end{minipage} & 50  & 918   \\
C8   &  Soybean-mintill                &\begin{minipage}[b]{0.08\columnwidth}
		\raisebox{-.45\height}{\includegraphics[width=\linewidth]{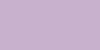}}
	\end{minipage} & 50  & 2418   \\
C9   &  Soybean-clean                  &\begin{minipage}[b]{0.08\columnwidth}
		\raisebox{-.45\height}{\includegraphics[width=\linewidth]{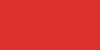}}
	\end{minipage} & 50  & 564  \\
C10   & Wheat                          &\begin{minipage}[b]{0.08\columnwidth}
		\raisebox{-.45\height}{\includegraphics[width=\linewidth]{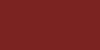}}
	\end{minipage} & 50  & 162  \\
C11   & Woods                          &\begin{minipage}[b]{0.08\columnwidth}
		\raisebox{-.45\height}{\includegraphics[width=\linewidth]{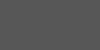}}
	\end{minipage} & 50  & 1244  \\
C12   & Buildings-Grass-Trees-Drives   &\begin{minipage}[b]{0.08\columnwidth}
		\raisebox{-.45\height}{\includegraphics[width=\linewidth]{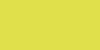}}
	\end{minipage} & 50  & 330  \\
C13   & Stone-Steel-Towers             &\begin{minipage}[b]{0.08\columnwidth}
		\raisebox{-.45\height}{\includegraphics[width=\linewidth]{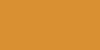}}
	\end{minipage} & 50  & 45  \\
C14   & Alfalfa                        &\begin{minipage}[b]{0.08\columnwidth}
		\raisebox{-.45\height}{\includegraphics[width=\linewidth]{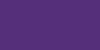}}
	\end{minipage} & 15  & 39  \\
C15   & Grass-pasture-mowed            &\begin{minipage}[b]{0.08\columnwidth}
		\raisebox{-.45\height}{\includegraphics[width=\linewidth]{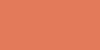}}
	\end{minipage} & 15  & 11  \\
C16   & Oats                           &\begin{minipage}[b]{0.08\columnwidth}
		\raisebox{-.45\height}{\includegraphics[width=\linewidth]{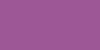}}
	\end{minipage} & 15  & 5  \\
 \hline\hline
Total     &                            & & 695 & 9671  \\
\bottomrule[1pt]
\end{tabular}
\end{center}
\end{table}

\begin{table}[t]
\begin{center}
\caption{Number of training and testing samples in Houston2013 data.}
\label{table:houston2013}
\begin{tabular}{ c | c c c c }
\toprule[1pt]
{Class}     &  {Class Name} & Color&  {Training} &  {Testing} \\
\hline\hline
C1   &  Grass-healthy   &  \begin{minipage}[b]{0.08\columnwidth}
		\raisebox{-.45\height}{\includegraphics[width=\linewidth]{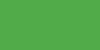}}
	\end{minipage}& 198  & 1053   \\
C2   &  Grass-stressed  &  \begin{minipage}[b]{0.08\columnwidth}
		\raisebox{-.45\height}{\includegraphics[width=\linewidth]{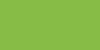}}
	\end{minipage}& 190  & 1064  \\
C3   &  Grass-synthetic &  \begin{minipage}[b]{0.08\columnwidth}
		\raisebox{-.45\height}{\includegraphics[width=\linewidth]{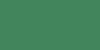}}
	\end{minipage}& 192  & 505   \\
C4   &  Tree            & \begin{minipage}[b]{0.08\columnwidth}
		\raisebox{-.45\height}{\includegraphics[width=\linewidth]{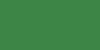}}
	\end{minipage} & 188  & 1056   \\
C5   &  Soil            &  \begin{minipage}[b]{0.08\columnwidth}
		\raisebox{-.45\height}{\includegraphics[width=\linewidth]{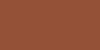}}
	\end{minipage}& 186  & 1056  \\
C6   &  Water           &  \begin{minipage}[b]{0.08\columnwidth}
		\raisebox{-.45\height}{\includegraphics[width=\linewidth]{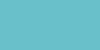}}
	\end{minipage}& 182  & 143   \\
C7   &  Residential     &  \begin{minipage}[b]{0.08\columnwidth}
		\raisebox{-.45\height}{\includegraphics[width=\linewidth]{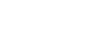}}
	\end{minipage}& 196  & 1072   \\
C8   &  Commercial      &  \begin{minipage}[b]{0.08\columnwidth}
		\raisebox{-.45\height}{\includegraphics[width=\linewidth]{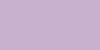}}
	\end{minipage}& 191  & 1053   \\
C9   &  Road            &  \begin{minipage}[b]{0.08\columnwidth}
		\raisebox{-.45\height}{\includegraphics[width=\linewidth]{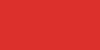}}
	\end{minipage}& 193  & 1059  \\
C10   & Highway         &  \begin{minipage}[b]{0.08\columnwidth}
		\raisebox{-.45\height}{\includegraphics[width=\linewidth]{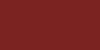}}
	\end{minipage}& 191  & 1036  \\
C11   & Railway         &  \begin{minipage}[b]{0.08\columnwidth}
		\raisebox{-.45\height}{\includegraphics[width=\linewidth]{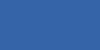}}
	\end{minipage}& 181  & 1054  \\
C12   & Parking-lot1    &  \begin{minipage}[b]{0.08\columnwidth}
		\raisebox{-.45\height}{\includegraphics[width=\linewidth]{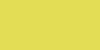}}
	\end{minipage}& 192  & 1041  \\
C13   & Parking-lot2    &  \begin{minipage}[b]{0.08\columnwidth}
		\raisebox{-.45\height}{\includegraphics[width=\linewidth]{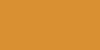}}
	\end{minipage}& 184  & 285  \\
C14   & Tennis-court    & \begin{minipage}[b]{0.08\columnwidth}
		\raisebox{-.45\height}{\includegraphics[width=\linewidth]{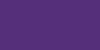}}
	\end{minipage} & 181  & 247  \\
C15   & Running-track   &  \begin{minipage}[b]{0.08\columnwidth}
		\raisebox{-.45\height}{\includegraphics[width=\linewidth]{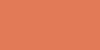}}
	\end{minipage}& 187  & 473 \\

 \hline\hline
Total     &             & & 2832 & 12197  \\
\bottomrule[1pt]
\end{tabular}
\end{center}
\end{table}

%

\subsection{Experimental Setups}\label{subsec:setups}

All the experiments in this paper are implemented under Pytorch 1.9.1, Cuda 11.2. The learning rate, epoch iteration, and training batch are set to 0.01, 500, and 64, respectively. The dimension of extracted features is set to 128. The structures of discriminative network in the experiments are set as 128-64-64-$\Lambda$ where $\Lambda$ denotes the number of pseudo classes.   If not specifice, $5\times 5$ neighbors is used to incorporate the spatial information. We adopt the stochastic gradient descent (SGD) as optimizer of deep model. The codes will be publicly available soon for easily replication at \url{https://github.com/shendu-sw/Adversarial_Learning_Intrinsic_Decomposition}.

\subsubsection{Evaluation Metrics}

We use the overall accuracy (OA), 
average accuracy (AA),
Kappa coefficient ($\kappa$) as measurements to evaluate performance. Furthermore, classification accuracy per class is also used to provide a thorough comparison. Besides, the visualization of classification maps is also provided to make a qualitative comparison.

\subsubsection{Baseline Methods}

Several representative baselines and backbone networks are selected for comparison. These methods denote the state-of-the-art CNNs (e.g.,  3-D CNN  \cite{b8},  PResNet \cite{b7},  HybridSN \cite{b6}), RNNs (e.g., RNN  \cite{b9}), GCNs (e.g., miniGCN \cite{b5}), and Transformers (e.g., ViT \cite{b1},  SpectralFormer \cite{b3},  SSFTTNet \cite{b4}), for hyperspectral image classification. 
\begin{itemize}
\item Support vector machine (SVM) is implemented through sklearn package and It is performed with radial basis function (RBF) kernel. SVM is chosen as the representative of such methods of non-machine learning.
\item The 3-D CNN  \cite{b8} consists of four subsequent convolutional block and each block accompanied with a ReLU activation function. Softmax layer and cross entropy classifier are finally added on the top layer of the 3-D CNN to classify different samples.
\item PResNet \cite{b7} is composed by several blocks of stacked convolutional layers, which have a bottleneck architecture (pyramidal bottleneck residual units) in which the output layer is larger than the input layer.
\item HybridSN was developed in \cite{b6}, and in the implementation, the architecture comprises of three three-dimensional convolution layers, one two-dimensional convolution layer and two fully connected layers. Each convolutional layer is accompanied with batch normalization and ReLU layer.
\item RNN \cite{b9} consists of two recurrent layers with gated recurrent unit (GRU) where each layer has 128 neural units.
\item The miniGCN follows the implementation in \cite{b5}, which successively contains
a BN layer, a graph convolutional layer with 128
neuron units, and a ReLU layer. 
\item The implementation of ViT \cite{b1} follow that in \cite{b3}. Only transformer encoders are used for classification task and five successive encoder blocks are used in the model's architecture.
\item As the former ViT, SpectralFormer \cite{b3} consists of five encoder blocks. Each encoder block consists of a four-head SA layer, a MLP with 8 hidden dimensions, and a GELU nonlinear activation layer. Specifically, the SpectralFormer contains the group-wise spectral embedding (GSE) and cross-layer adaptive fusion (CAF).
\item SSFTTNet \cite{b4} adopts the architecture as the released code at \url{https://github.com/zgr6010/HSI_SSFTT}.
\end{itemize}

\subsection{Evaluation of the Computational Performance}
At first, we test the computational performance of the proposed method compared with other methods. 
In this set of experiments, the HybridSN is chosen as the backbone CNN to extract features. 
In order to demonstrate the general usability of the proposed method, a common machine with a Intel@Xeon(R) Gold 6226R GPU, 128GB RAM and Quadro RTX 6000 24GB GPU is used to evaluate the classification performance. The training and testing cost of 3-D CNN, PResNet, HybridSN, SpectralFormer are selected for comparison. 

Table \ref{table:computational} shows the computational performance over the three datasets. From the table, we can find that the training of proposed AdverDecom took about 618.9s, 141.5s, and 382.6s over Pavia University data, Indian Pines data, and Houston2013 data, respectively. The proposed method took a comparable computational efficiency when compared with 3-D CNN and HybridSN while presented a better computational efficiency than PResNet and SpectralFormer. Furthermore, the testing of the proposed AdverDecom cost about 1.87s, 0.58s, and 0.61s separately which could satisfy the computational efficiency requirements of most applications.

\begin{table}[t]
\begin{center}
\caption{Computational performance over different datasets.}
\label{table:computational}
\begin{tabular}{ c | c || c c c}
\toprule[1pt]
{Data}     & {Metrics} &  {Pavia University} &  Indian Pines&  Houston2013  \\
\hline\hline
  \multirow{2}{*}{3-D CNN}  &  Training(s)        &  471.24 & 112.7&308.9  \\
  &  Testing(s)      &  1.225 & 0.45&0.425  \\
\hline
 \multirow{2}{*}{PResNet}&  Training(s)  &  1752.4 & 364.0&961.24  \\
  &  Testing(s)        &  7.1 & 2.1&2.2  \\
\hline 
\multirow{2}{*}{HybridSN}  &  Training(s)      &  536.2 & 123.9&320.5 \\
   &  Testing(s)  &  1.7 & 0.56&0.55  \\
\hline 
\multirow{2}{*}{SpectralFormer}  & Training(s)      &  1061.2 & 232.9&621.3  \\
   &  Testing(s)  &  2.52 & 0.92&0.77  \\
\hline 
\multirow{2}{*}{AdverDecom}  &  Training(s)      &  618.9 & 141.5&382.6    \\
   &  Testing(s)  &  1.87 & 0.58&0.61   \\
\bottomrule[1pt]
\end{tabular}
\end{center}
\end{table}

\subsection{Evaluation of the Models Trained with Different Backbone CNNs}
  The backbone CNN influences the quality of extracted environmental-related and category-related features, and thus shows significant effect on the classification performance of the hyperspectral image.
In this set of experiments, we test the performance of the proposed method with different backbone CNNs, s.t., 3-D CNN, PResNet, and HybridSN.  The structures of these backbone CNNs are set as the setups in subsection \ref{subsec:setups}.

Table \ref{table:pavia_cnns}, \ref{table:indian_cnns}, and \ref{table:houston2013_cnns} shows the comparison results of the proposed method and the Vanilla CNNs over the three datasets.
Inspect the comparison results in these tables and it can be noted that the following hold.

First, the performance based on PResNet and HybridSN is better than that based on 3-D CNN. For Pavia University data, the proposed method can obtain 93.94\%, 94.13\% with PResNet and HybridSN as backbone which is better than that with 3-D CNN (88.64\%).  For Indian Pines data, the proposed method can obtain 88.03\%, 88.50\%, 91.07\% with 3-D CNN, PResNet, HybridSN as backbones separately. As for Houston2013, the proposed method can obtain 86.30\%, 88.60\%, and 90.03\% with 3-D CNN, PResNet, HybridSN as backbones, respectively.
Then, the proposed deep intrinsic decomposition with adversarial learning can remarkably improve the performance of vanilla CNN.
For Pavia University, the proposed method can improve the performance by 1.12\%, 3.83\%, 3.86\% with 3-D CNN, PResNet, HybridSN as backbone model, respectively.
As for Indian Pines, the proposed method can obtain an improvement by 10.81\%, 5.53\%, 12.35\% with the three different backbones. While for Houston2013 data, the proposed method can impprove the performance by 1.59\%, 3.01\%, and 3.14\%, respectively.

\begin{table}[t]
\begin{center}
\caption{Classification accuracies (OA, AA, and $\kappa$) of the proposed method with different backbone over Pavia University data.}
\label{table:pavia_cnns}
\begin{tabular}{ c | c || c c c}
\toprule[1pt]
\multirow{2}{*}{Data}     &  \multirow{2}{*}{Metrics} &  \multicolumn{3}{c}{CNN Backbone}  \\
\cline{3-5}
     &   &  {3-D CNN}&  {PResNet}&  {HybridSN}   \\
\hline\hline
   &  OA(\%)        &  87.52 & 90.11&90.27   \\
 Vanilla  &  AA(\%)      &  89.01 & 89.43&91.79    \\
   &  $\kappa$(\%)  &  83.37 & 86.68&87.03    \\
\hline
   &  OA(\%)        &  88.64 & 93.94&94.13   \\
 Proposed  &  AA(\%)      &  82.34 & 93.19&93.63    \\
   &  $\kappa$(\%)  &  84.51 & 91.82&92.11    \\
\bottomrule[1pt]
\end{tabular}
\end{center}
\end{table}

\begin{table}[t]
\begin{center}
\caption{Classification accuracies (OA, AA, and $\kappa$) of the proposed method with different backbone over Indian Pines data.}
\label{table:indian_cnns}
\begin{tabular}{ c | c || c c c}
\toprule[1pt]
\multirow{2}{*}{Data}     &  \multirow{2}{*}{Metrics} &  \multicolumn{3}{c}{CNN Backbone}  \\
\cline{3-5}
     &   &  {3-D CNN}&  {PResNet}&  {HybridSN}   \\
\hline\hline
   &  OA(\%)        &  77.22 & 82.97&78.72   \\
 Vanilla  &  AA(\%)      &  86.83 & 90.19&88.15    \\
   &  $\kappa$(\%)  &  74.21 & 80.65&75.81    \\
\hline
   &  OA(\%)        &  {88.03} &88.50 &91.07    \\
  Proposed &  AA(\%)      &  92.09 &92.94 &{95.45}    \\
   &  $\kappa$(\%)  &  {86.30} &86.82 &89.79    \\
\bottomrule[1pt]
\end{tabular}
\end{center}
\end{table}

\begin{table}[t]
\begin{center}
\caption{Classification accuracies (OA, AA, and $\kappa$) of the proposed method with different backbone over Houston2013 data.}
\label{table:houston2013_cnns}
\begin{tabular}{ c | c || c c c}
\toprule[1pt]
\multirow{2}{*}{Data}     &  \multirow{2}{*}{Metrics} &  \multicolumn{3}{c}{CNN Backbone}  \\
\cline{3-5}
     &   &  {3-D CNN}&  {PResNet}&  {HybridSN}   \\
\hline\hline
   &  OA(\%)        &  {84.71} &85.59 &86.89    \\
  Vanilla &  AA(\%)      &  85.53 &87.45 &{88.92}    \\
   &  $\kappa$(\%)  &  {83.40} &84.35 &85.77    \\
\hline
   &  OA(\%)       & 86.30  &88.60 &90.03    \\
  Proposed &  AA(\%)     & 88.34  &89.91 &91.66    \\
   &  $\kappa$(\%) & 85.12  &87.63 &89.18    \\
\bottomrule[1pt]
\end{tabular}
\end{center}
\end{table}

\subsection{Evaluation of the Models Trained with Different Number of Pseudo Classes $K$}

The construction of pseudo classes is an important factor for the learning of the discriminative network and therefore, it can also influence the classification performance. The number of pseudo classes defines the class of environmental factors. When $K$ is set to 1, it means that all the samples possess the same environmental factor.In the experiments, the $k$ is set to $\{1,2,3,4,5,6,7,8,9,10,20,30\}$. Table \ref{table:pseudo_class} presents the results of the proposed method with different number of pseudo classes over the three datasets, respectively. 

From the figure, we can conclude that a proper $K$ can guarantee a good performance of the proposed method. For Pavia Unviersity data, the performance achieve the best when the $K$ is set to 5. For Indian Pines data, the performance achieve an accuracy of 91.07\% OA which is the best when $K$ is set to 2. While for Houston2013 data, the performance achieve the best (90.03\%) when $K$ is set to 4.
Even though the proposed method performs different with different $K$, one has a large range to select $K$ since some $K$ performs similar. 
That is, within a certain range, the $K$ is not sensitive to the classification performance.
For example, for Pavia University data, the proposed method can achieve similar performance when $K$ is set to 3 (93.58\%), 4(93.91\%), 5(94.13\%), 6(93.37\%), 7(93\%). If there is a specific requirement for high accuracy, one can use cross validation for a proper $K$.

\begin{table*}[t]
\begin{center}
\caption{Classification accuracies (OA, AA, and $\kappa$) of the proposed method with different number of pseudo classes.}
\label{table:pseudo_class}
\begin{tabular}{ c | c || c c c c c c c c c c c c c}
\toprule[1pt]
\multirow{2}{*}{Data}     &  \multirow{2}{*}{Metrics} &  \multicolumn{12}{c}{Number of Pseudo Classes}  \\
\cline{3-14}
     &   &  { 1}&  { 2}&  { 3} &  {4} & {5} & {6} & {7} & {8} & {9} & 10 & 20 & 30  \\
\hline\hline
   &  OA(\%)        &  91.34 &92.49 &93.58 &93.91 &{\bf 94.13} &93.37 &93.00 &92.87 &92.34 &  93.48& 93.74 &  93.97\\
 PU  &  AA(\%)      &  90.95 &91.95 &92.05 &93.59 &{\bf 93.63} &93.39 &92.77 &92.73 &92.64  & 93.51 & 92.50 & 93.28 \\
   &  $\kappa$(\%)  &  88.32 &89.88 &91.31 &91.82 &{\bf 92.11} &91.10 &90.61 &90.39 &89.72   & 91.20 & 91.57 & 91.90\\
\hline
   &  OA(\%)       & 88.34 & {\bf 91.07}  &91.01 &90.88 &89.65 &90.51 &87.67 &88.83 & 90.07 & 90.20 & 87.17 & 84.23 \\
  IP &  AA(\%)     & 94.30 & {\bf 95.45}  &94.92 &95.09 &94.05 &95.42 & 92.93& 94.13& 94.73  & 94.74 & 93.33 & 91.75 \\
   &  $\kappa$(\%) & 86.70 & {\bf 89.79}  &89.69 &89.57 &88.13 &89.17 &85.91 &87.25 & 88.67 & 88.78 & 85.36 & 81.98 \\
\hline
   &  OA(\%)       &  86.55  & 86.14 &87.01 &{\bf 90.03} & 89.60 & 89.42 &89.08 &88.93 &88.86   & 89.11 & 86.06 & 86.09 \\
  HS &  AA(\%)     &  88.61  & 88.47&89.15 & {\bf 91.66}& 91.24& 91.05& 90.62& 90.72  & 90.83 & 90.26 & 87.84& 88.27\\
   &  $\kappa$(\%) &  85.40  & 84.96&85.89 & {\bf 89.18}& 88.71 & 88.51& 88.15& 87.99&  87.90  & 88.18 & 84.87 & 84.90\\
\bottomrule[1pt]
\end{tabular}
\end{center}
\end{table*}

\subsection{Evaluation of the Models Trained with Different $\alpha$}

\begin{figure*}[t]
\centering
 \subfigure[]{\label{subfig:pavia}\includegraphics[width=0.32\linewidth]{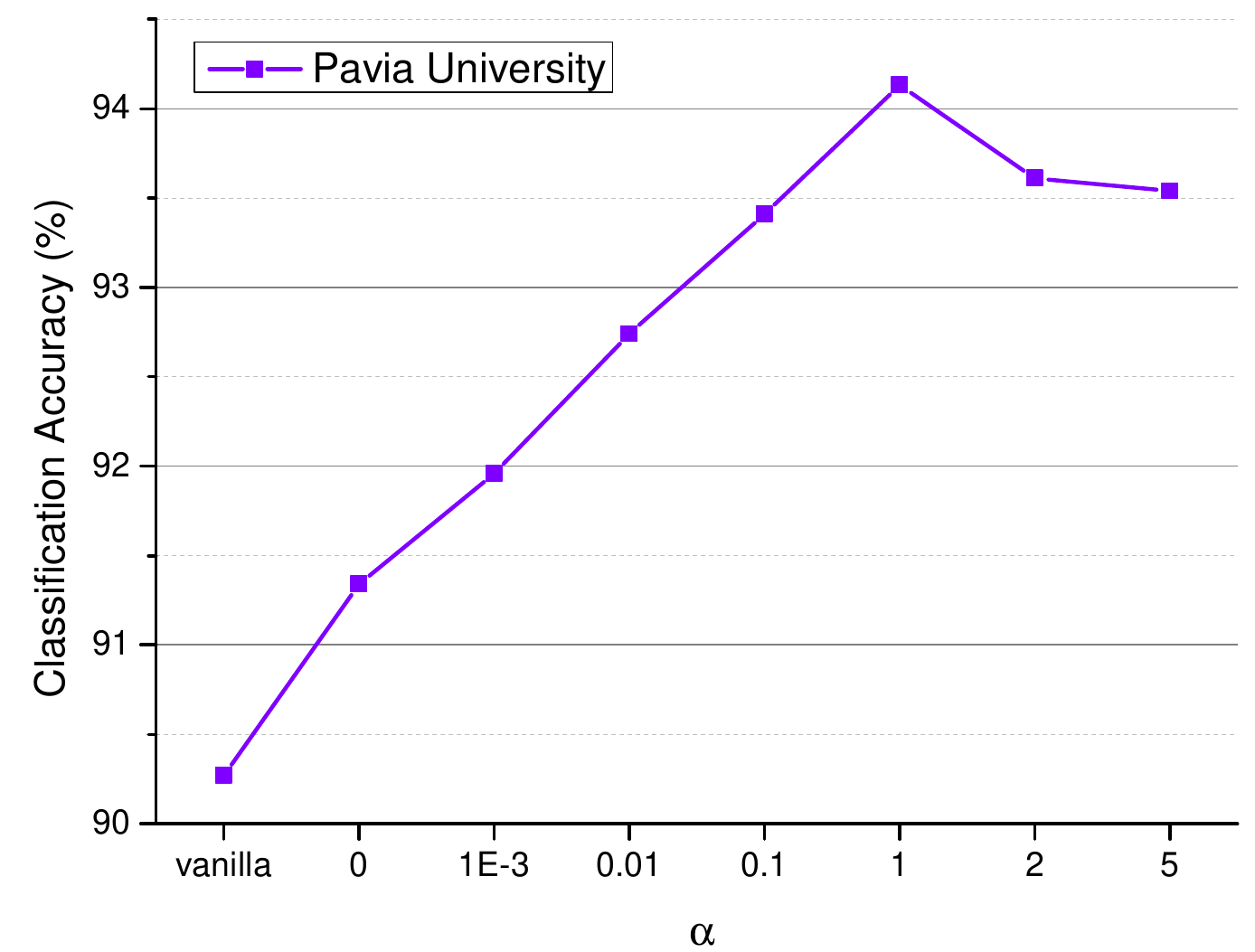}}
 \subfigure[]{\label{subfig:pavia_label}\includegraphics[width=0.32\linewidth]{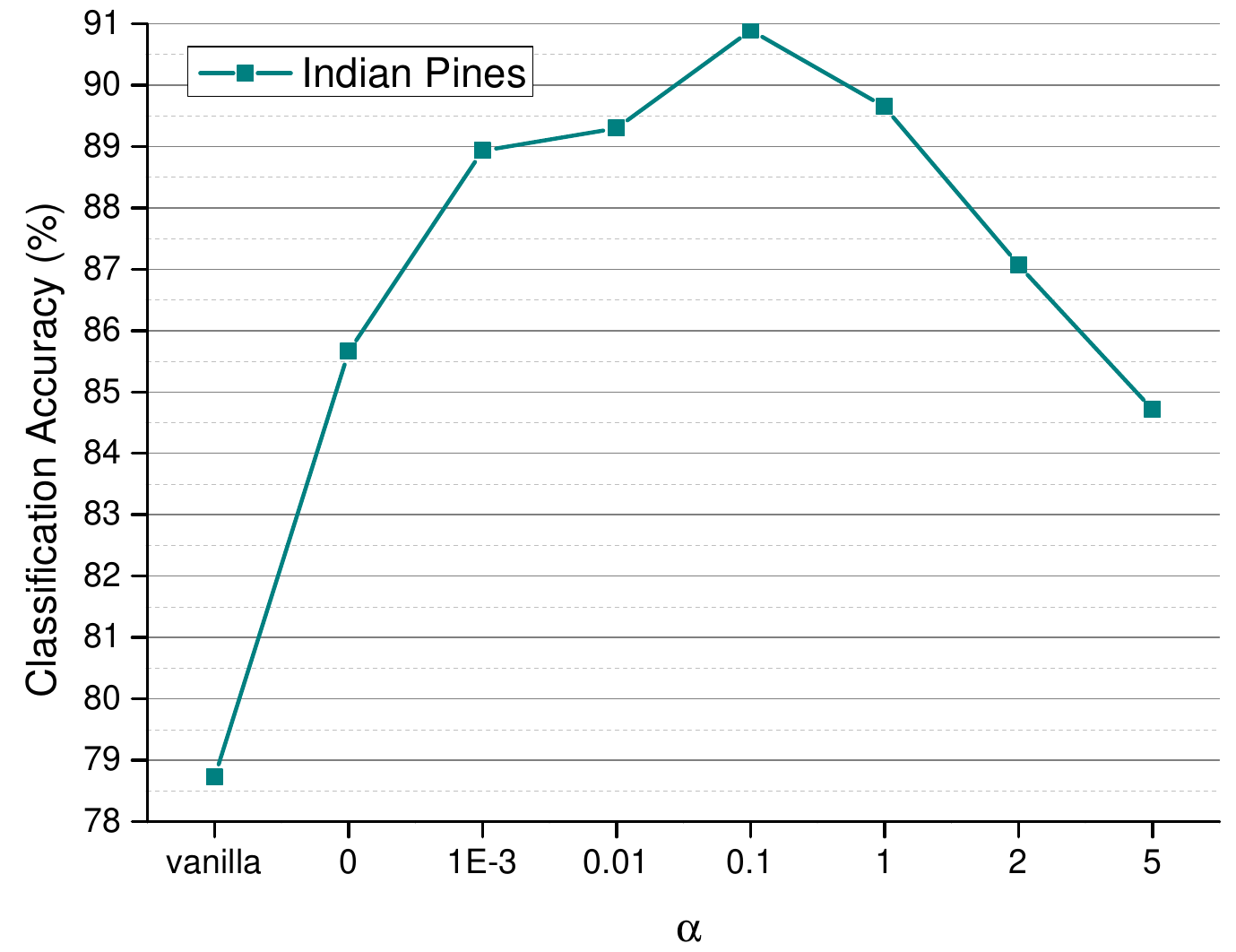}}
 \subfigure[]{\label{subfig:mapping}\includegraphics[width=0.32\linewidth]{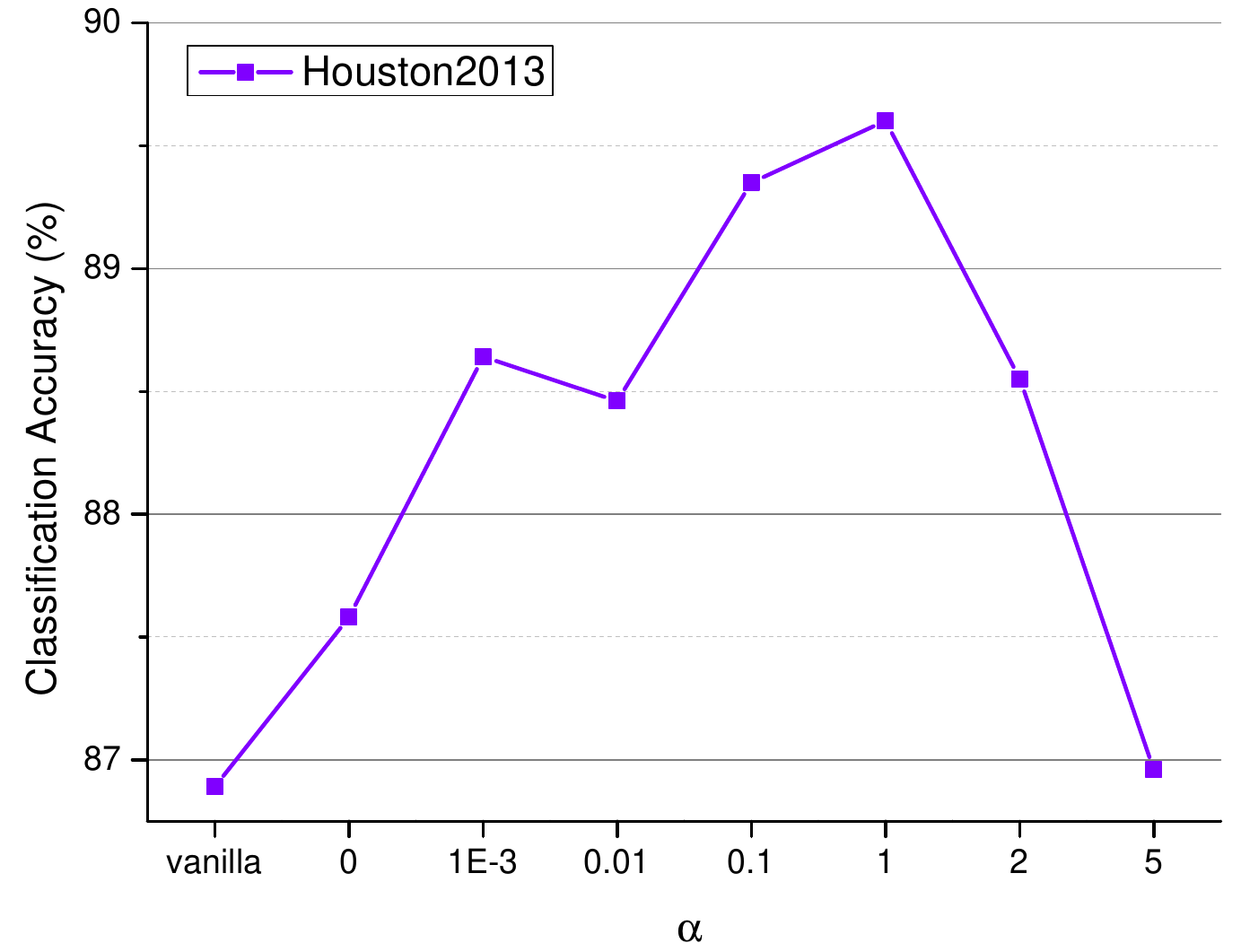}}
   \caption{Classification performance with different $\lambda$ over (a) Pavia University; (b) Indian Pines; (c) Houston2013.}
\label{fig:lambda}
\end{figure*}

As mentioned in Section \ref{sec:proposed}, $\alpha$ denotes the tradeoff between the adversarial error and classification error. It can significantly affect the learning process of environmental-related features and category-related features, and thus influence the classification performance.
Generally, a larger $\alpha$ value leads to a better performance. However, excessively large $\alpha$ values decrease the classification performance and even bring about the non-converge of the deep model.
The reason is that a larger $\alpha$ means a higher weight for adversarial learning and reduces the excessive intra-class variation caused by environmental factors. 
As a result, the learned features can be easily to discriminate different classes and thus increase the classification performance.
However, excessively large $\alpha$ focuses on too much attention on the environmental-related features and ignores the category-related features, which in turn decreases the classification performance.
Fig. \ref{fig:lambda} shows the tendiencies of the performance with different $\alpha$ over the three datasets.

Here, we choose the value of $\alpha$ from $\{0, 0.001, 0.01, 0.1, 1, 2, 5\}$. It should be noted that when we conduct the experiments when $\alpha$ is set to 10, the deep model cannot converge over the three datasets. 
As the figure shows, a larger $\alpha$ provide a better performance while excessively $\alpha$ values decrease the performanc. Since $\alpha$ value can significantly affect the performance of the model, a proper $\alpha$ is essential for current task. Cross-validation can be used to choose a proper $\alpha$ faced with different tasks.

Besides, we can conclude from Fig. \ref{fig:lambda} that over Pavia University data and Houston2013 data, the proposed method can achieve the best when $\alpha$ is set to 1. While over Indian Pines data, the proposed method can achieve 90.90\% which performs the best when $\alpha$ is set to 0.1.

\subsection{Evaluation of the Models Trained with Different Size of Spatial Neighbors}

\begin{figure*}[t]
\centering
 \subfigure[]{\label{subfig:pavia}\includegraphics[width=0.32\linewidth]{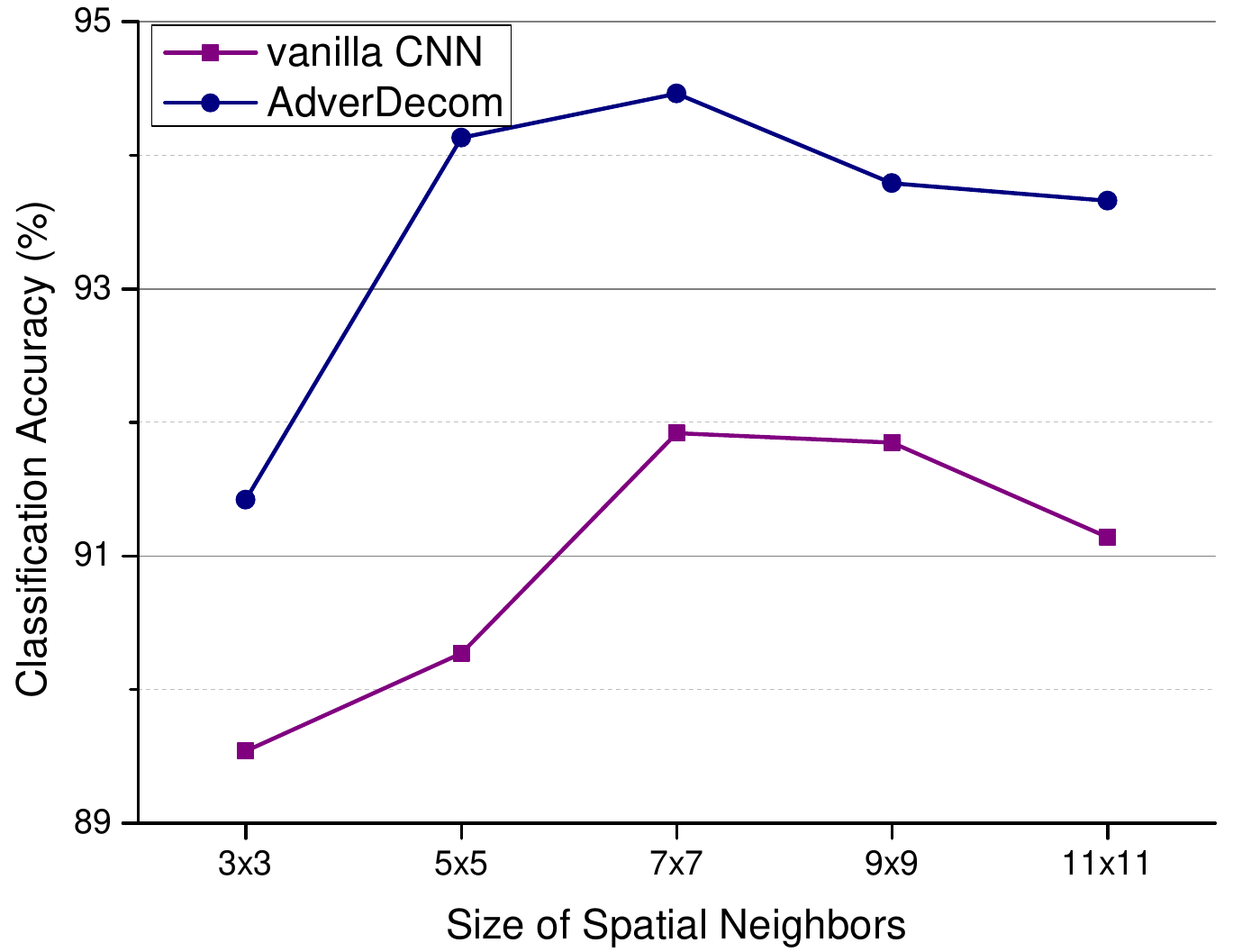}}
 \subfigure[]{\label{subfig:pavia_label}\includegraphics[width=0.32\linewidth]{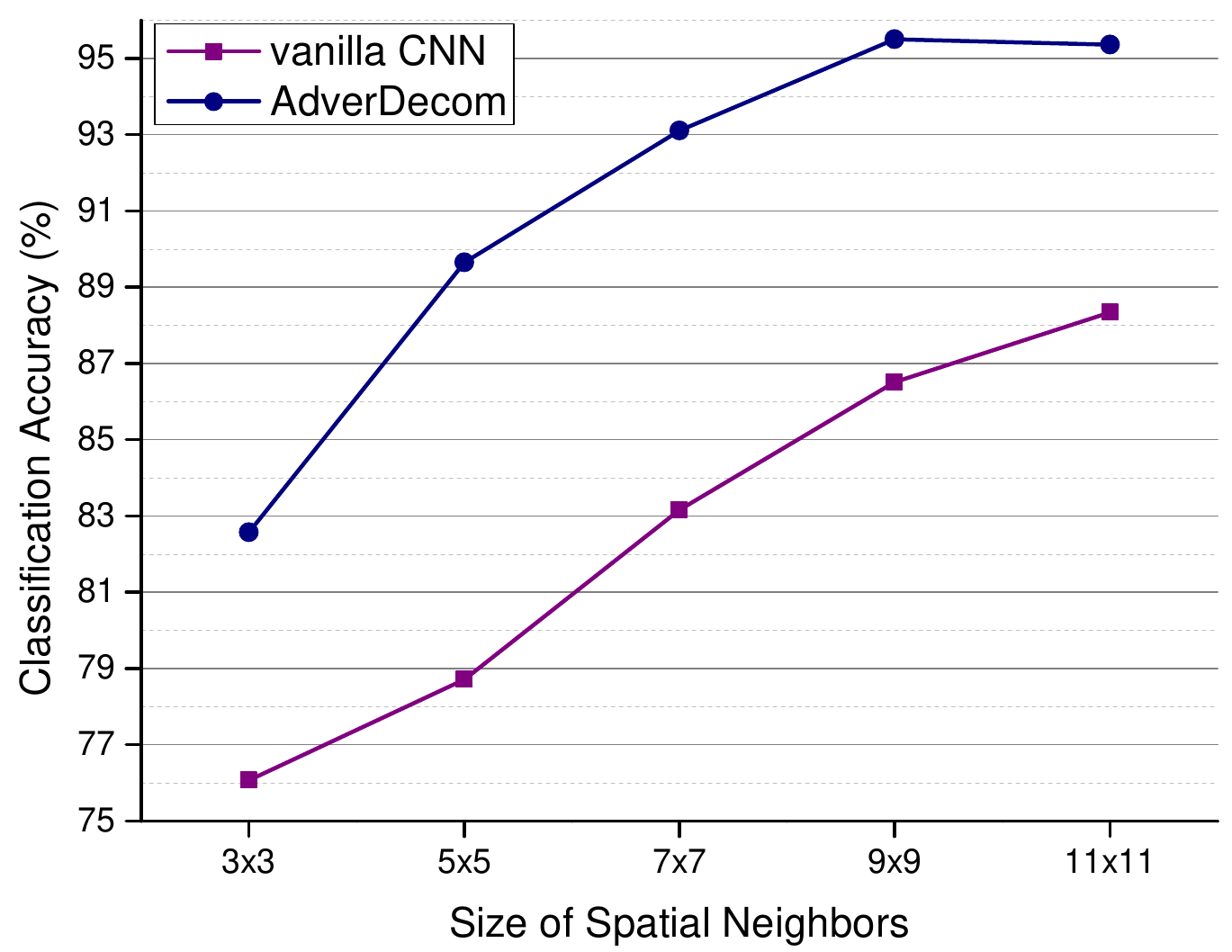}}
 \subfigure[]{\label{subfig:mapping}\includegraphics[width=0.32\linewidth]{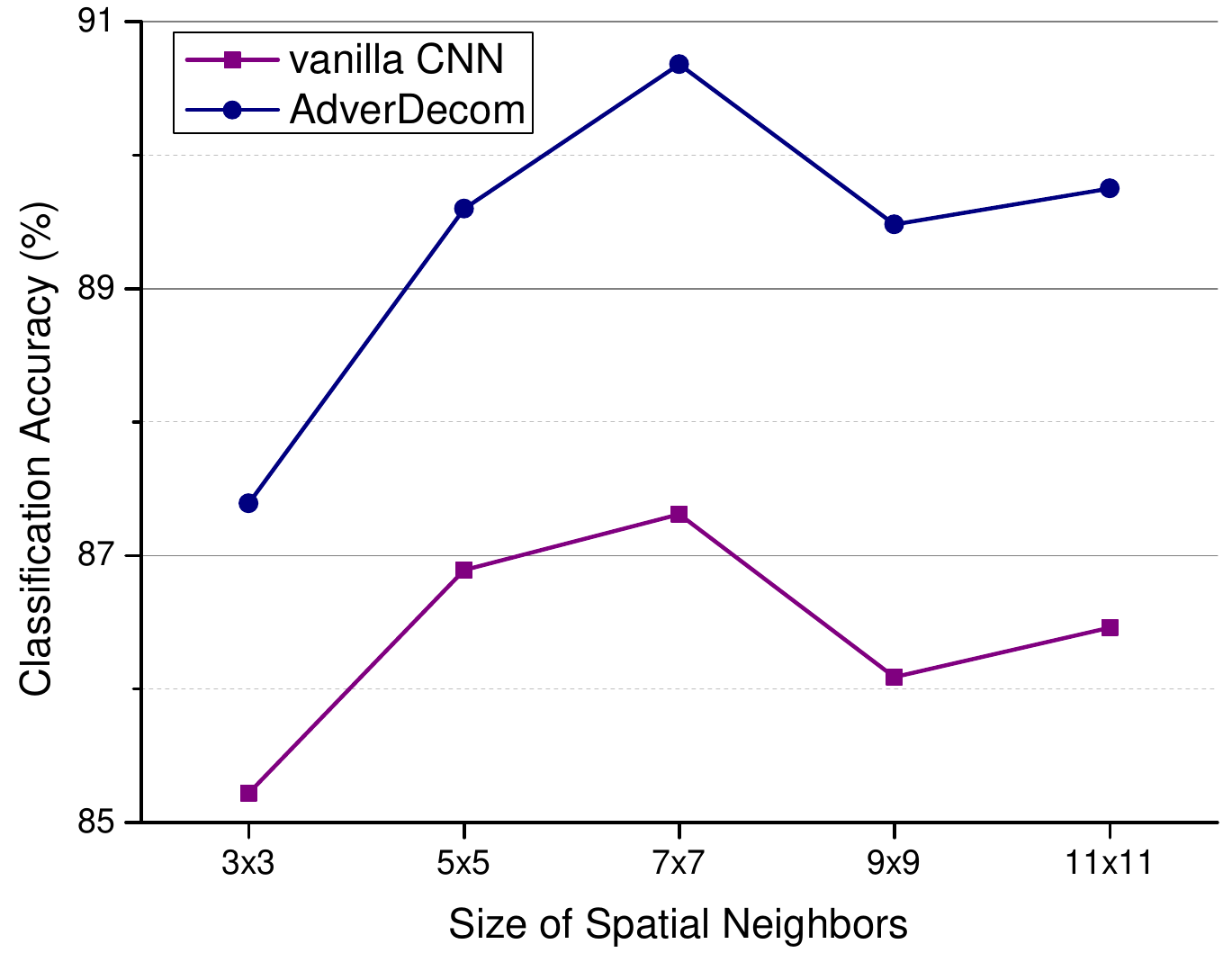}}
   \caption{Classification performance with different size of spatial neighbors over (a) Pavia University; (b) Indian Pines; (c) Houston2013.}
\label{fig:size}
\end{figure*}

It is obvious that the size of spatial neighbors can significantly affect the classification performance of the hyerspectral images. Therefore, in this subsection, we further investigate the effects of the neighbor size on the classification performance. The neighbor sizes are chosen from \{$3\times 3$, $5\times 5$, $7\times 7$, $9\times 9$, $11\times 11$\}. Fig. \ref{fig:size} shows the tendencies of classification accuracies under different neighbor sizes.

As shown in Fig. \ref{fig:size}, the classification accuracy of the proposed method can provide an improvement of performance under different size of neighbors.
For Pavia University data, the proposed method achieves the best (94.46\%) with $7\times 7$ size of neighbors and we can obtain a 2.54\% improvement when compared with vanilla CNN. For Houston2013, the proposed method also achieve the best performance (90.68\%) under $7\times 7$ neighbors.
As for Indian Pines data, the performance increases with the increase of the neighbor size and the accuracy can obtain 95.36\% with $11\times 11$ neighbor size. 
Generally, samples with larger neighbor sizes contain more spatial information and thus can provide a better classification performance, just as Indian Pines data. However, larger neighbor sizes imply a more complex physical model, which increases the difficulty of model training. Therefore, for Pavia University and Houston2013 data, samples with $7\times 7$ neighbors can provide a better classification accuracy than samples with $11\times 11$ neighbors.

\subsection{Evaluation of the Models Trained with Different Number of Samples}

\begin{figure*}[t]
\centering
 \subfigure[]{\label{subfig:pavia}\includegraphics[width=0.32\linewidth]{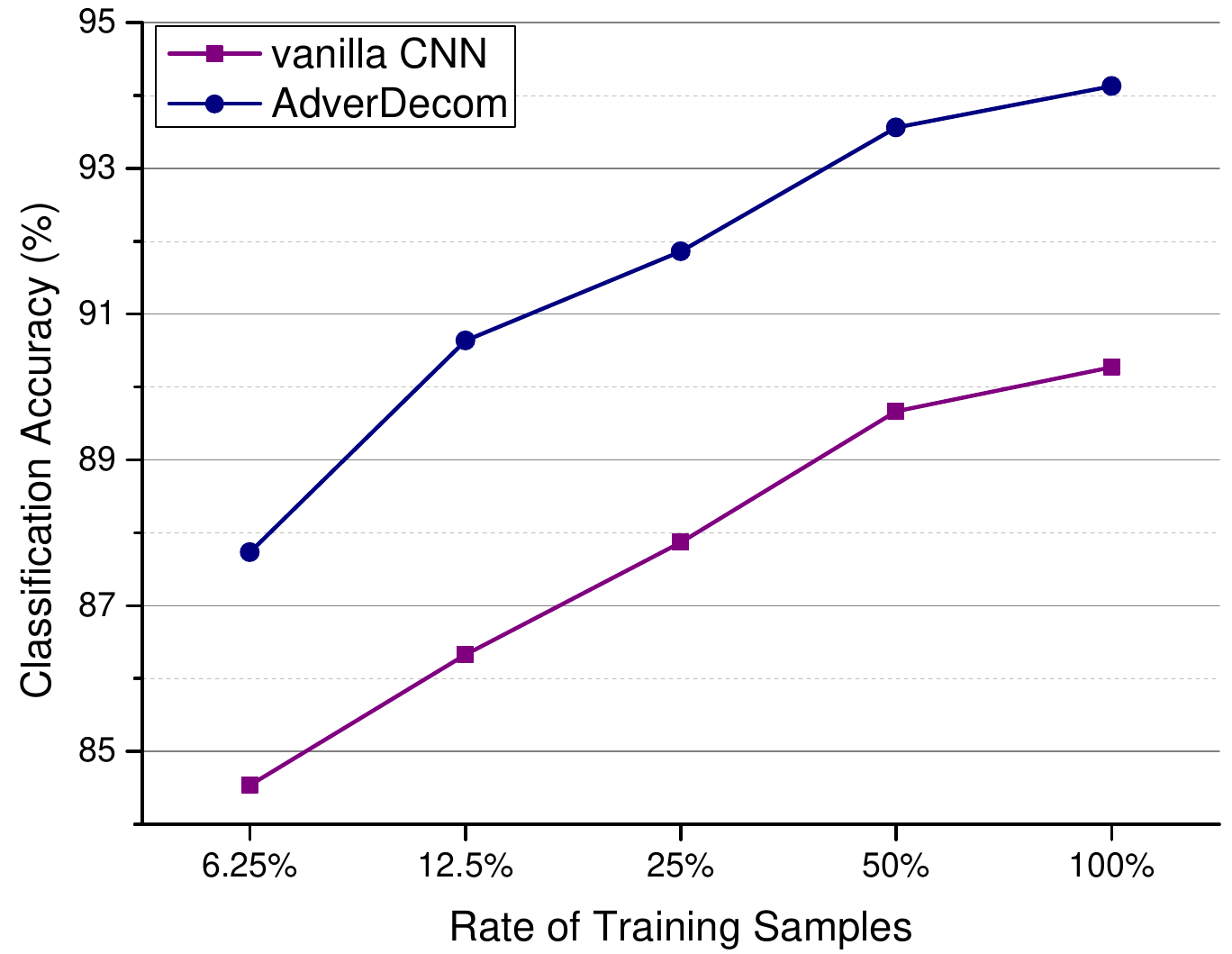}}
 \subfigure[]{\label{subfig:pavia_label}\includegraphics[width=0.32\linewidth]{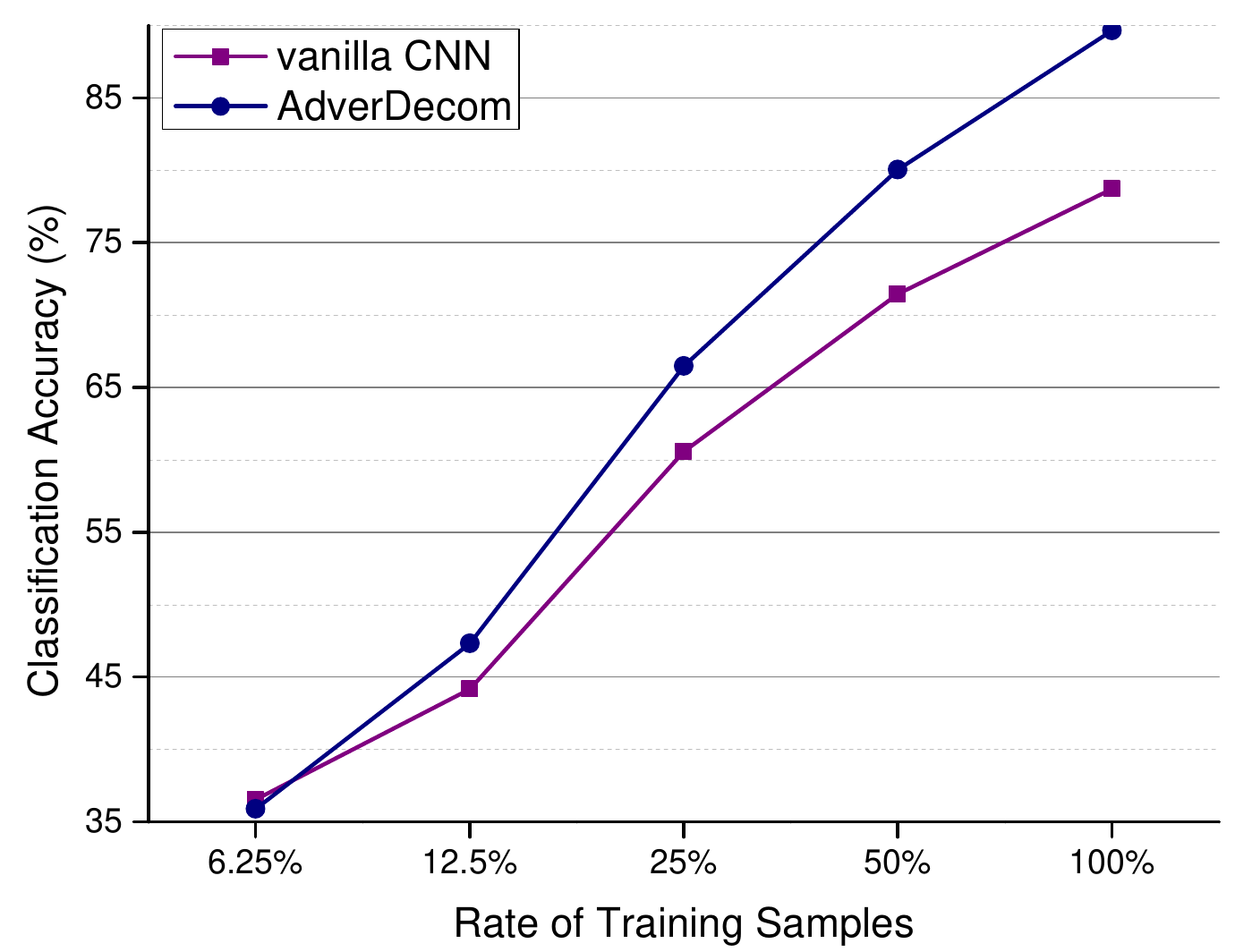}}
 \subfigure[]{\label{subfig:mapping}\includegraphics[width=0.32\linewidth]{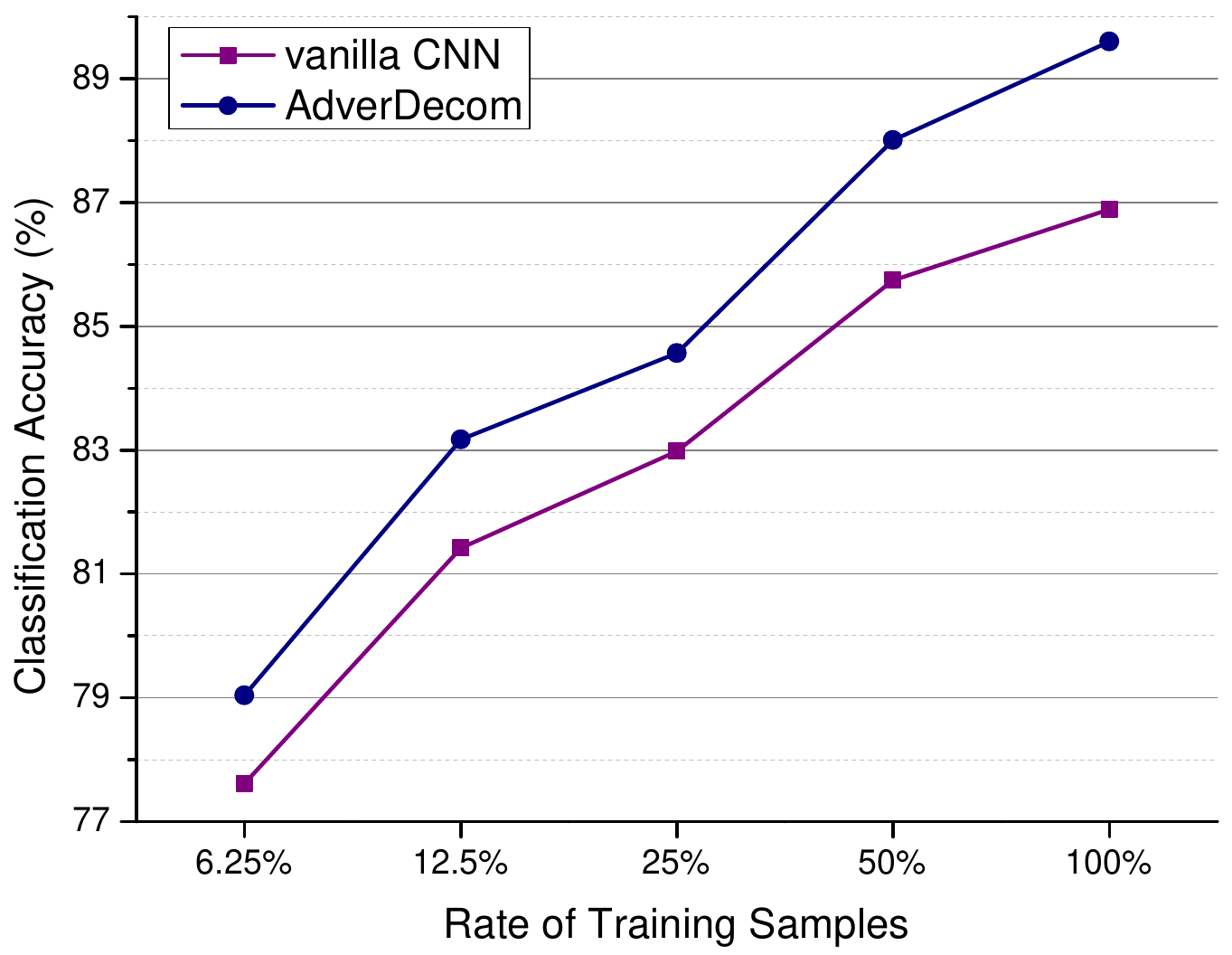}}
   \caption{Classification performance with different rate of training samples over (a) Pavia University; (b) Indian Pines; (c) Houston2013.}
\label{fig:number}
\end{figure*}

Prior subsections mainly conduct the experiments over a given training and testing samples divided as Table \ref{table:pavia}, \ref{table:indian}, and \ref{table:houston2013} list.
This subsection will further evaluate the performance of the developed method under a different number of training samples. 
As shows in Table \ref{table:pavia}-\ref{table:houston2013}, 3921 training and 40002 testing samples for Pavia University data, 695 training and 9671 testing samples for Indian Pines data, 2832 training and 12197 testing samples for Houston2013 data, are used for experiments, respectively. While in this set of experiments, 6.25\%, 12.5\%, 25\%, 50\%, and 100\% samples are selected from the original training samples over these datasets to evaluate the performance with different number of training samples. That is, over Pavia University data, 245, 490, 980, 1960, 3921 training samples are selected. Over Indian Pines data, 43, 86, 173, 347, 695 samples are selected and 177, 354, 708, 1416, 2832 samples are chosen for Houston2013 data.
Fig. \ref{fig:number} shows the tendencies of classification performance with different number of training samples over the three datasets, respectively.

We can find that the accuracies by the proposed method can be remarkably improved compared with the vanilla CNN. For Pavia University data, the accuracy can be increased by about 3\%-4\%. For Houston2013 data, the accuracy can be increased by about 1.5\%-3\%. Specifically, for Indian Pines data, the accuracy can be even increased by more than 10\%.
This is because the proposed method decomposes the environmental-related features and the category-related features, and improves the discrimination of category-related features and reduces the impact of environmental factors on hyperspectral image classification.
Besides, the classification performance of the learned model is significantly improved with the increase of training samples. More 
training samples provides additional information for the deep model to learn, allowing it to better extract discriminative features for hyerspectral image classification.

Furthermore, we show the classification maps over different datasets in Figs. \ref{fig:pavia}-\ref{fig:houston2013} under setups in Table \ref{table:pavia}-\ref{table:houston2013}, respectively. Compare Fig. \ref{subfig:pavia_6} with \ref{subfig:pavia_12}, \ref{subfig:indian_6} with \ref{subfig:indian_12}, and \ref{subfig:houston2013_6} with \ref{subfig:houston2013_12}, and we can find that the classification error can be significantly decreased under the proposed AdverDecom method. This also indicates that the proposed method with deep intrinsic decomposition through adversarial learning can provide a more discriminative feature by decompose the environmental-related and category-related features.

\subsection{Comparison with State-of-the-art Methods}

To further validate the effectiveness of the proposed method for hyperspectral image classification, we compare classification results of the proposed method with the state-of-the-art methods. Table \ref{table:pavia_comparison}, \ref{table:indian_comparison}, and \ref{table:houston2013_comparison} present the comparisons over the three datasets, respectively. All the experimental results in these tables come from the same experimental setups.

From Table \ref{table:pavia_comparison}, we can obtain that the proposed method can obtain 94.13\% OA that outperforms the CNNs (e.g. 3-D CNN (87.52\%), PResNet (90.11\%), HybridSN (90.27\%)), RNN (80.61\%), miniGCN (83.23\%), and Transformers (e.g. ViT (86.27\%), SpectralFormer (90.04\%), SSFTTNet(82.56\%)) over Pavia University data. As listed in Table \ref{table:indian_comparison}, for Indian Pines data, the proposed method can provide an accuracy of 91.07\% outperforms that of the CNNs (e.g. 3-D CNN (77.22\%), PResNet (82.97\%), HybridSN (78.72\%)), RNN (81.11\%), miniGCN (74.71\%), and Transformers (e.g. ViT (65.16\%), SpectralFormer (83.38\%), SSFTTNet (80.29\%)). Furthermore, for Houston2013 data, the proposed method can also provide a better classification performance when compared with other state-of-the-art methods (see Table \ref{table:houston2013_comparison} for details).   These comparison results show the effectiveness of the proposed method for current task. 

\begin{figure*}[t]
\centering
 \subfigure[]{\label{subfig:pavia_1}\includegraphics[width=0.16\linewidth]{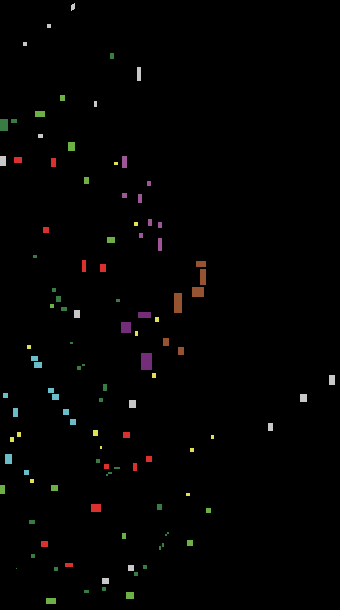}}
 \subfigure[]{\label{subfig:pavia_2}\includegraphics[width=0.16\linewidth]{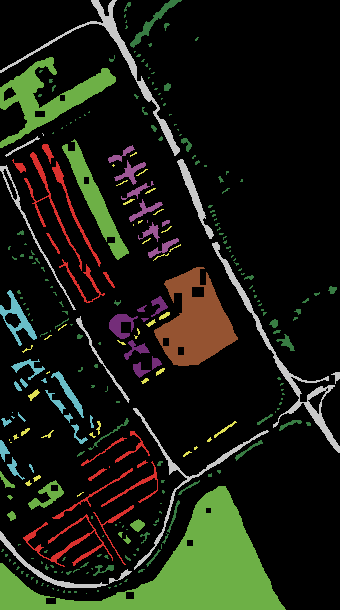}}
 \subfigure[]{\label{subfig:pavia_3}\includegraphics[width=0.16\linewidth]{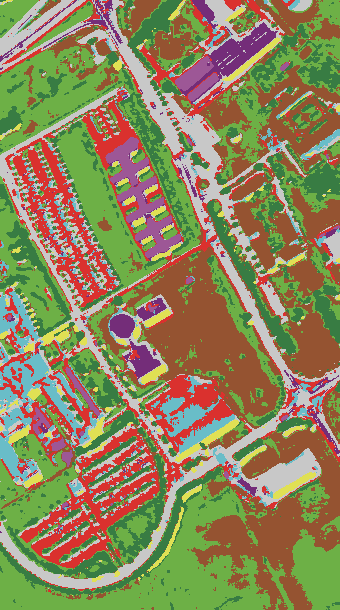}}
\subfigure[]{\label{subfig:pavia_4}\includegraphics[width=0.16\linewidth]{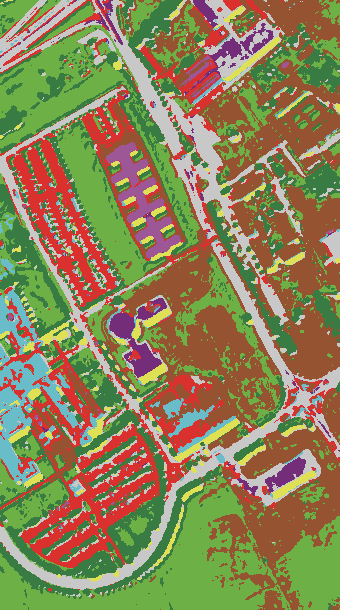}}
\subfigure[]{\label{subfig:pavia_5}\includegraphics[width=0.16\linewidth]{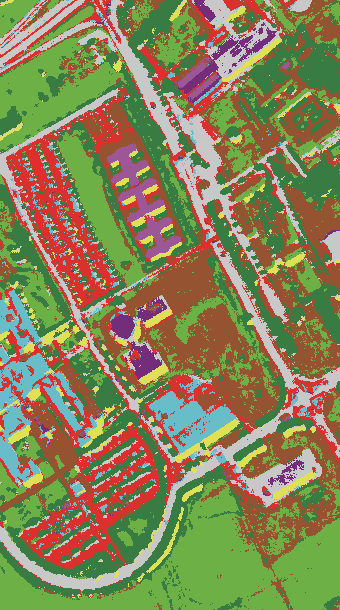}}
\subfigure[]{\label{subfig:pavia_6}\includegraphics[width=0.16\linewidth]{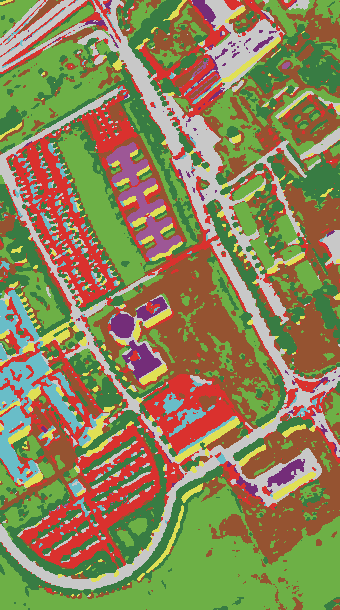}}
\subfigure[]{\label{subfig:pavia_7}\includegraphics[width=0.16\linewidth]{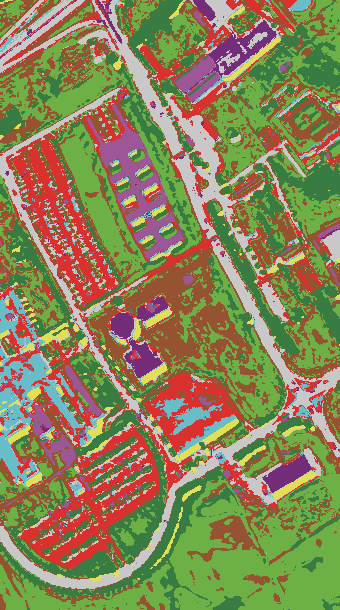}}
\subfigure[]{\label{subfig:pavia_8}\includegraphics[width=0.16\linewidth]{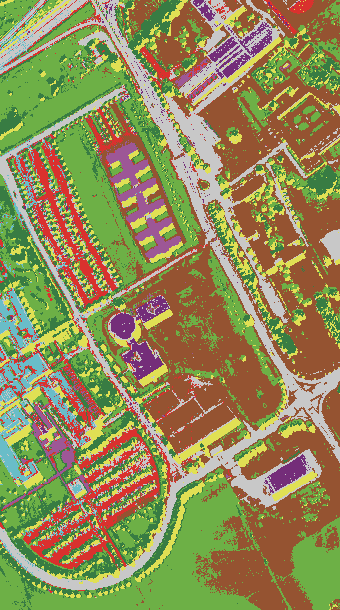}}
\subfigure[]{\label{subfig:pavia_9}\includegraphics[width=0.16\linewidth]{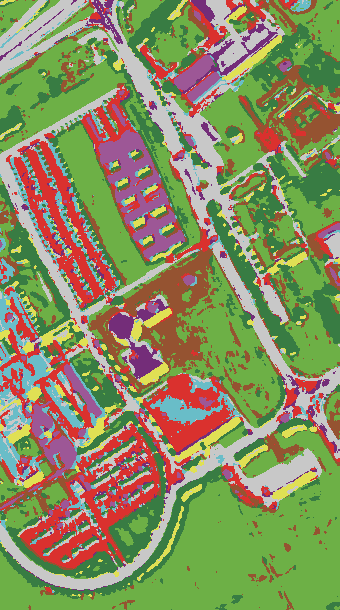}}
\subfigure[]{\label{subfig:pavia_10}\includegraphics[width=0.16\linewidth]{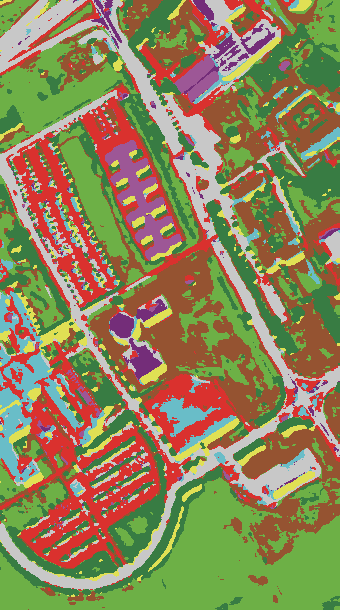}}
\subfigure[]{\label{subfig:pavia_11}\includegraphics[width=0.16\linewidth]{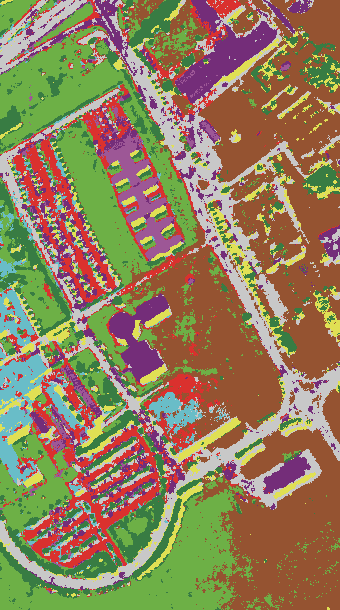}}
\subfigure[]{\label{subfig:pavia_12}\includegraphics[width=0.16\linewidth]{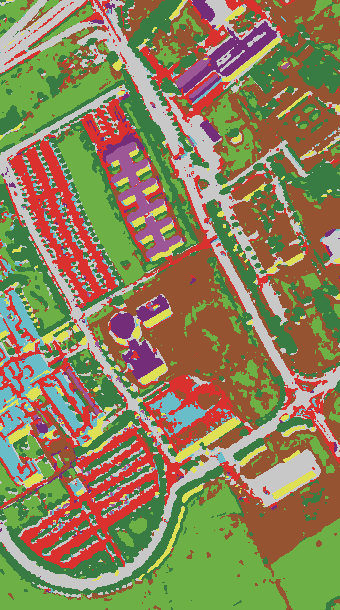}}
   \caption{Pavia University data. (a) Training; (b) Testing; (c) SVM; (d) 3-D CNN; (e) PResNet; (f) HybridSN; (g) RNN; (h) miniGCN; (i) ViT; (j) SpectralFormer; (k) SSFTTNet; (l) AdverDecom.}
\label{fig:pavia}
\end{figure*}

\begin{figure*}[t]
\centering
 \subfigure[]{\label{subfig:indian_1}\includegraphics[width=0.16\linewidth]{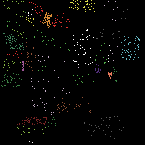}}
 \subfigure[]{\label{subfig:indian_2}\includegraphics[width=0.16\linewidth]{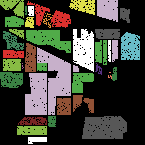}}
 \subfigure[]{\label{subfig:indian_3}\includegraphics[width=0.16\linewidth]{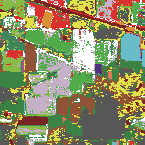}}
\subfigure[]{\label{subfig:indian_4}\includegraphics[width=0.16\linewidth]{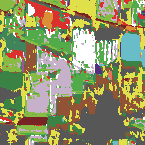}}
\subfigure[]{\label{subfig:indian_5}\includegraphics[width=0.16\linewidth]{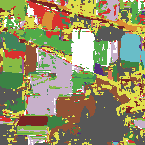}}
\subfigure[]{\label{subfig:indian_6}\includegraphics[width=0.16\linewidth]{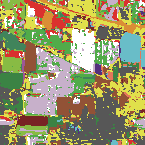}}
\subfigure[]{\label{subfig:indian_7}\includegraphics[width=0.16\linewidth]{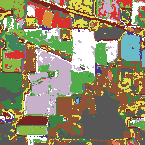}}
\subfigure[]{\label{subfig:indian_8}\includegraphics[width=0.16\linewidth]{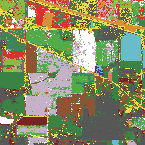}}
\subfigure[]{\label{subfig:indian_9}\includegraphics[width=0.16\linewidth]{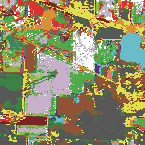}}
\subfigure[]{\label{subfig:indian_10}\includegraphics[width=0.16\linewidth]{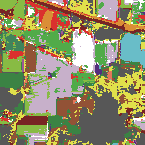}}
\subfigure[]{\label{subfig:indian_11}\includegraphics[width=0.16\linewidth]{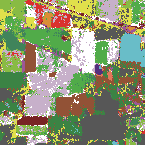}}
\subfigure[]{\label{subfig:indian_12}\includegraphics[width=0.16\linewidth]{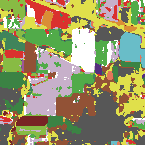}}

   \caption{Indian Pines data. (a) Training; (b) Testing; (c) SVM; (d) 3-D CNN; (e) PResNet; (f) HybridSN; (g) RNN; (h) miniGCN; (i) ViT; (j) SpectralFormer; (k) SSFTTNet; (l) AdverDecom.}
\label{fig:indian}
\end{figure*}

\begin{figure*}[t]
\centering
 \subfigure[]{\label{subfig:houston2013_1}\includegraphics[width=0.49\linewidth]{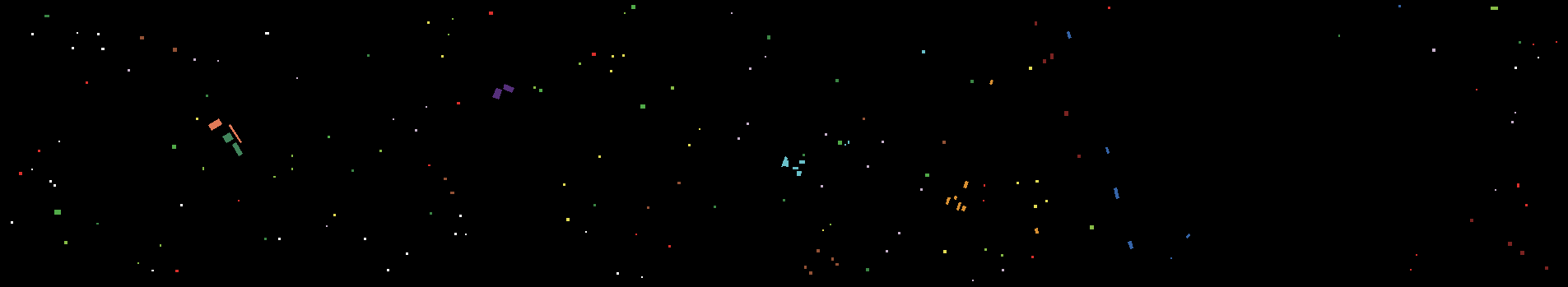}}
 \subfigure[]{\label{subfig:houston2013_2}\includegraphics[width=0.49\linewidth]{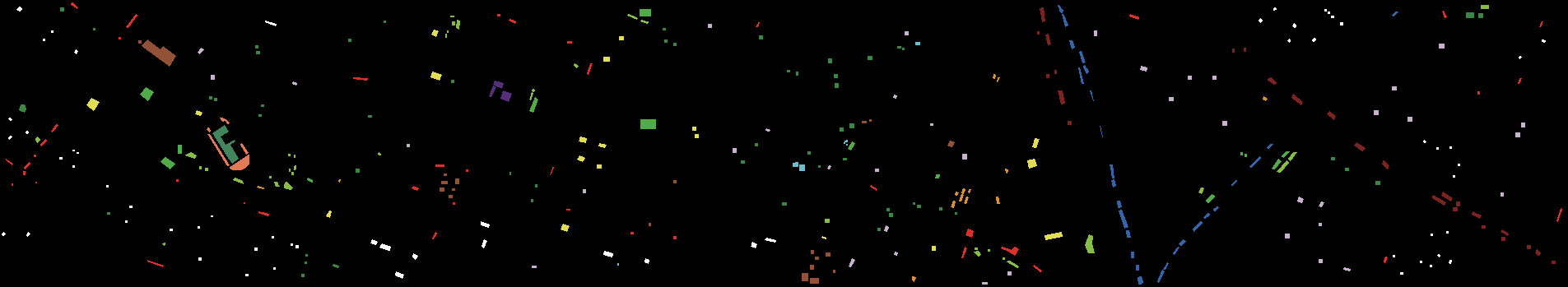}}
 \subfigure[]{\label{subfig:houston2013_3}\includegraphics[width=0.49\linewidth]{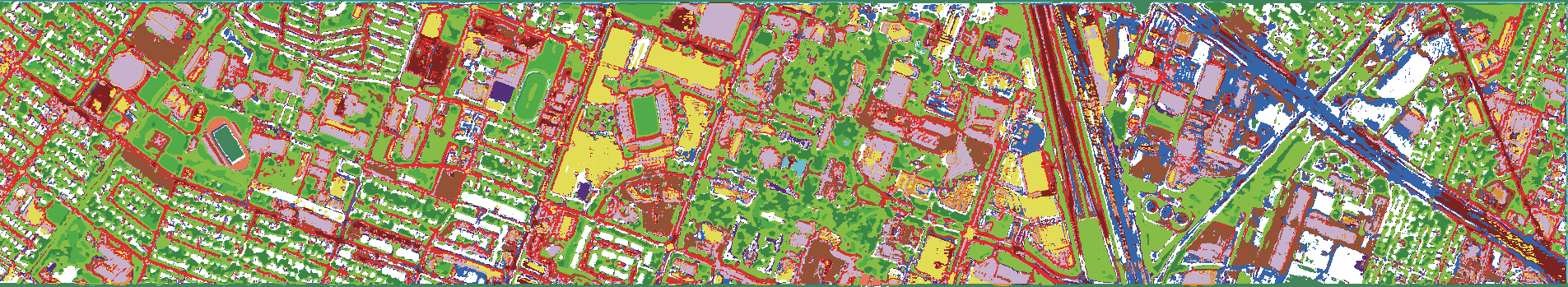}}
\subfigure[]{\label{subfig:houston2013_4}\includegraphics[width=0.49\linewidth]{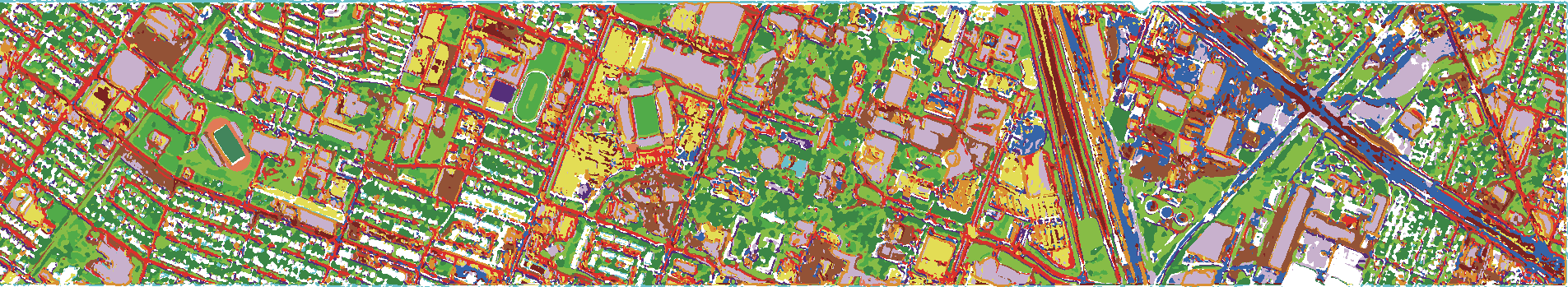}}
\subfigure[]{\label{subfig:houston2013_5}\includegraphics[width=0.49\linewidth]{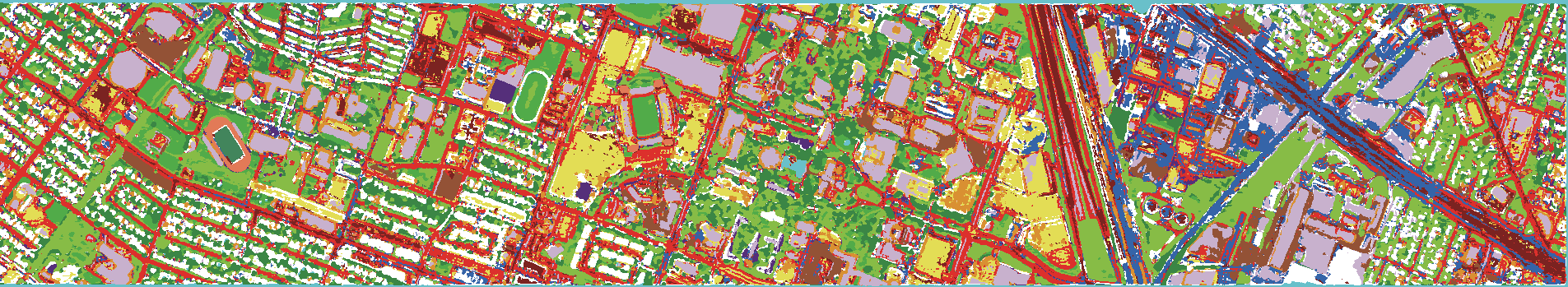}}
\subfigure[]{\label{subfig:houston2013_6}\includegraphics[width=0.49\linewidth]{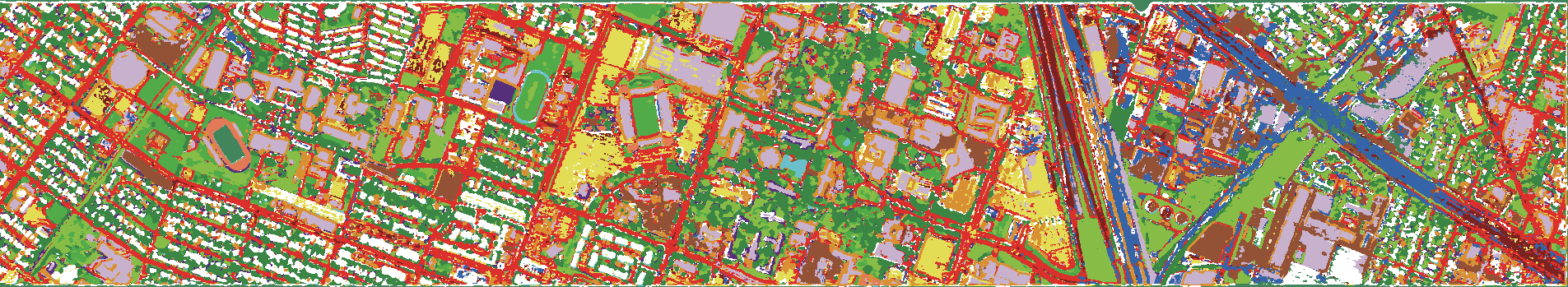}}
\subfigure[]{\label{subfig:houston2013_7}\includegraphics[width=0.49\linewidth]{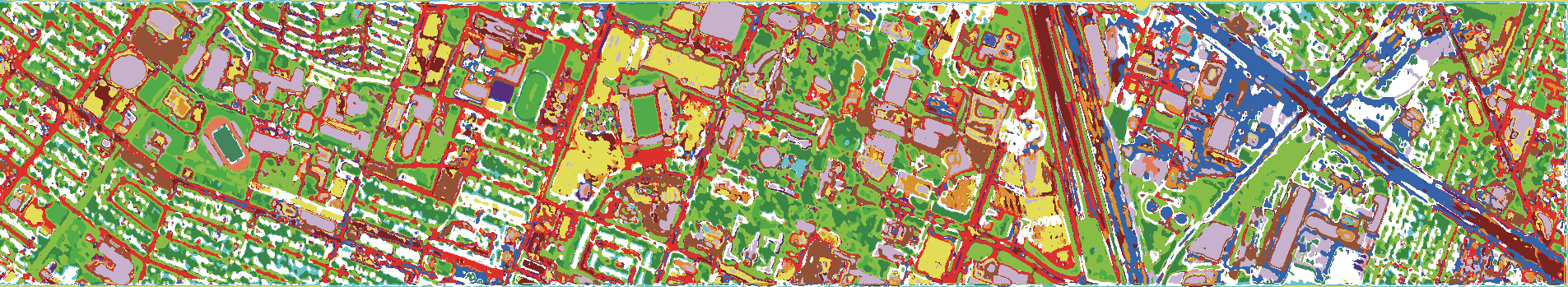}}
\subfigure[]{\label{subfig:houston2013_8}\includegraphics[width=0.49\linewidth]{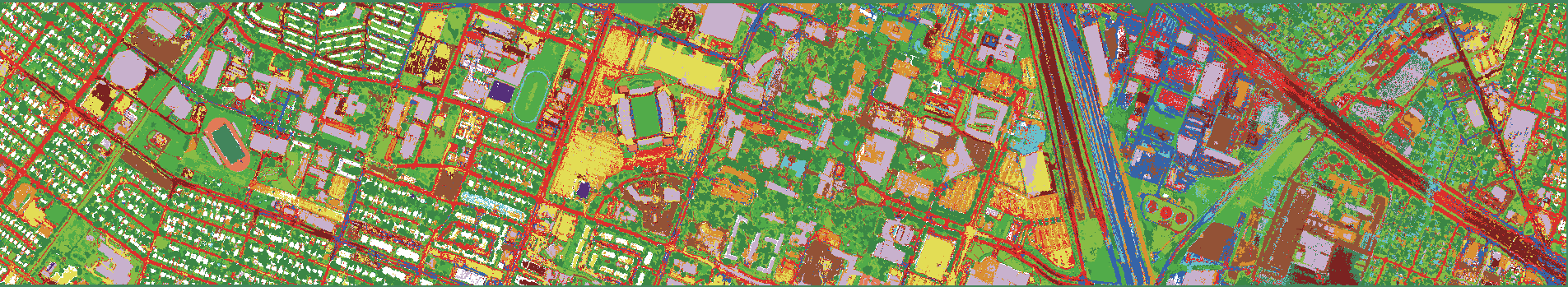}}
\subfigure[]{\label{subfig:houston2013_9}\includegraphics[width=0.49\linewidth]{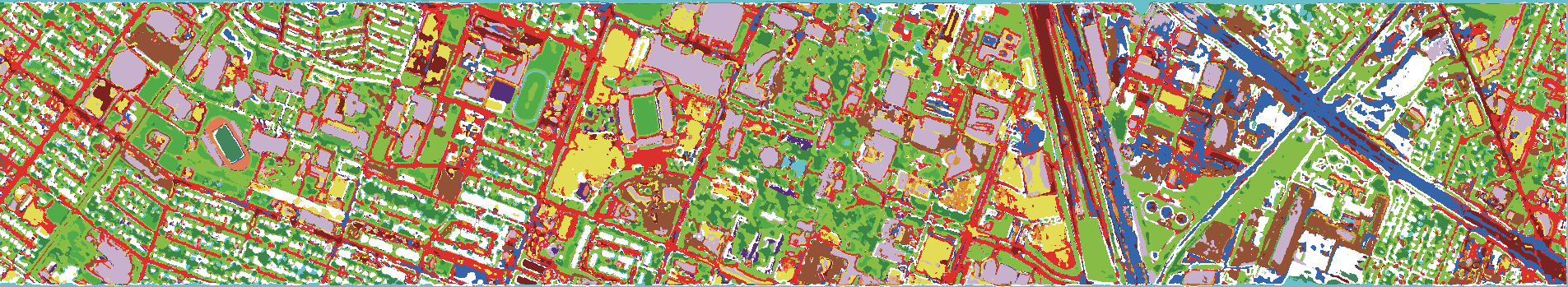}}
\subfigure[]{\label{subfig:houston2013_10}\includegraphics[width=0.49\linewidth]{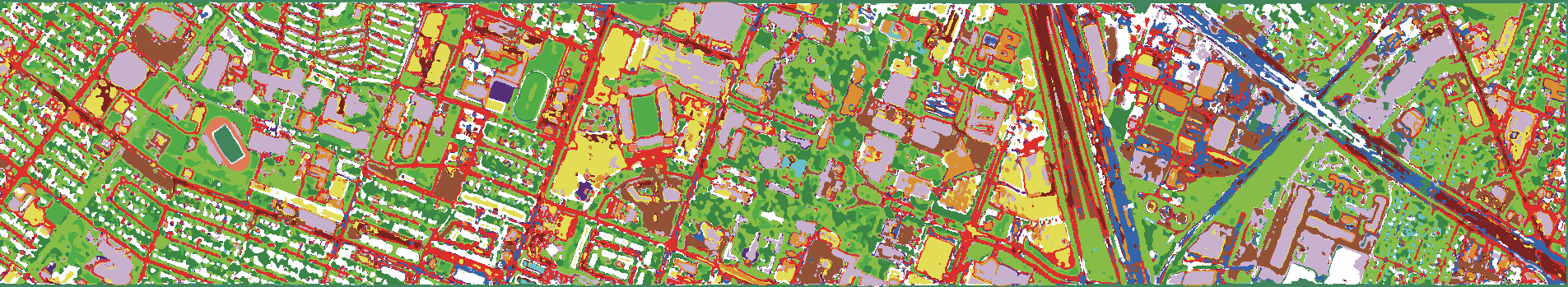}}
\subfigure[]{\label{subfig:houston2013_11}\includegraphics[width=0.49\linewidth]{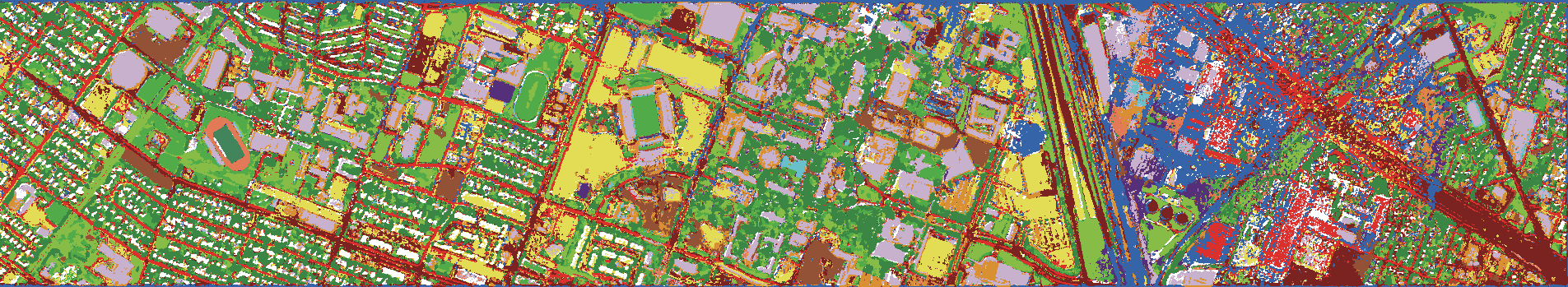}}
\subfigure[]{\label{subfig:houston2013_12}\includegraphics[width=0.49\linewidth]{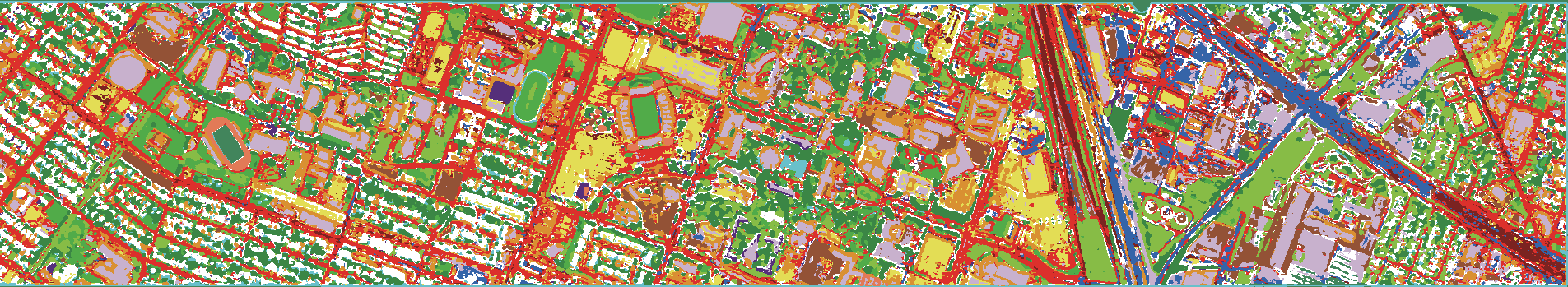}}

   \caption{Houston2013 data. (a) Training; (b) Testing; (c) SVM; (d) 3-D CNN; (e) PResNet; (f) HybridSN; (g) RNN; (h) miniGCN; (i) ViT; (j) SpectralFormer; (k) SSFTTNet; (l) AdverDecom.}
\label{fig:houston2013}
\end{figure*}

Besides, from classification maps in Figs \ref{fig:pavia}-\ref{fig:houston2013}, we can also find that the classification error of the proposed method can be decreased by the proposed AdverDecom and thus the accuracy can be obviously improved. In particular, the results of our proposed methods  have less noisy points compared to other state-of-the-art methods.

To sum up,  the proposed method can significantly improve the representational ability of the deep model and significantly improve the classification accuracy when compared with not only other handcrafted methods and CNNs-based deep models, but also other state-of-the-art deep methods.

\begin{table*}[t]
\setlength{\aboverulesep}{-0.1pt}
\setlength{\belowrulesep}{-0.1pt}
\begin{center}
\caption{Classification accuracies (OA, AA, and $\kappa$) of different methods achieved on the Pavia University data.}
\label{table:pavia_comparison}
\begin{tabular}{ c || c | c c c | c | c | c c c || c}
\toprule[1pt]
\multirow{2}{*}{Methods}     &  \multirow{2}{*}{SVM} &  \multicolumn{3}{c|}{CNNs} &  \multirow{2}{*}{RNN} & \multirow{2}{*}{miniGCN} & \multicolumn{3}{c||}{Transformers} &\multirow{2}{*}{AdverDecom} \\
 \cmidrule(lr){3-5} \cmidrule(lr){8-10}
    &   &  { 3-D CNN}&  {PResNet }&  { HybridSN} &   &  & { ViT} & {SpectralFormer} & SSFTTNet & \\
\hline\hline
 C1   &  77.08  & 82.95  &80.11 &83.49 &85.56 & 91.55 &82.55 &84.72 &75.89 & {\bf 92.43}\\
 C2   &  79.22  & 88.32  &94.81 &91.52 &75.00 & 84.62 &{\bf 96.57} &95.86 &81.46 & 95.32\\
 C3   &  77.52  & 73.55  &{\bf 85.62} &81.76 &71.63 & 74.27 &56.42 &66.72 &75.54 & 80.22\\
 C4   &  94.61  & 93.96  &98.49 &{\bf 98.80} &94.16 & 71.22 &95.98 &96.53 &83.72 & 98.73\\
 C5   &  98.74  & 99.55  &99.91 &100.0 &91.46 & 99.55 &93.62 &99.19 &{\bf 100.0} & 98.83\\
 C6   &  {\bf 93.68}  & 79.11  &79.48 &85.39 &72.66 & 67.61 &49.43 &73.16 &83.44 & 89.57\\
 C7   &  85.12  & 90.21  &79.00 &91.85 &92.25 & 86.75 &79.61 &79.71 &{\bf 99.80} & 91.64\\
 C8   &  93.82  & 98.63  &91.77 &94.41 &95.54 & 86.15 &93.79 &97.74 &88.61 & {\bf 99.20}\\
 C9   &  92.58  & 94.84  &95.72 &98.87 &93.21 & {\bf 100.0} &91.07 &93.33 &95.60 & 96.73\\
                                                                                                             
 \hline\hline
 {OA(\%)}      & 83.76  & 87.52 & 90.11 &90.27 &80.61 &83.23 &86.27 &90.04  & 82.56 & {\bf 94.13}\\
 
 {AA(\%)}      & 88.04  & 89.01 & 89.43 &91.79 &85.72 &84.64 &82.12 &87.44  & 87.12 & {\bf 93.63}\\
 
 {$\kappa$(\%)} & 78.98  & 83.37 & 86.68 &87.03 &74.83 &77.44 &81.16 &86.51  &77.30  & {\bf 92.11}\\
\bottomrule[1pt]
\end{tabular}
\end{center}
\end{table*}

\begin{table*}[t]
\setlength{\aboverulesep}{-0.1pt}
\setlength{\belowrulesep}{-0.1pt}
\begin{center}
\caption{Classification accuracies (OA, AA, and $\kappa$) of different methods achieved on the Indian Pines data. }
\label{table:indian_comparison}
\begin{tabular}{ c || c | c c c | c | c | c c c || c}
\toprule[1pt]
\multirow{2}{*}{Methods}     &  \multirow{2}{*}{SVM} &  \multicolumn{3}{c|}{CNNs} &  \multirow{2}{*}{RNN} & \multirow{2}{*}{miniGCN} & \multicolumn{3}{c||}{Transformers} &\multirow{2}{*}{AdverDecom} \\
 \cmidrule(lr){3-5} \cmidrule(lr){8-10}
    &   &  { 3-D CNN}&  {PResNet }&  { HybridSN} &   &  & { ViT} & {SpectralFormer} & SSFTTNet & \\
\hline\hline
 C1   & 72.11 &67.85     &72.76 & 66.04&72.76 &70.52  &59.68 &78.97 &{83.37} & {\bf 86.71} \\
 C2   & 71.43 &77.04     &87.50 & 79.46&84.82 &53.19  &37.76 &85.08 &71.39 &{\bf 93.88} \\
 C3   & 86.96 &93.48     &94.02 & {94.57}&78.26 &91.85  &55.43 &85.33 &94.57 &{\bf 97.28} \\
 C4   & 95.97 &92.84     &{ 94.85} & 91.28&88.14 &93.74  &65.10 &94.18 &93.93 &{\bf 98.43} \\
 C5   & 88.67 &83.21     &91.97 & 90.96&83.50 &95.12  &86.51 &84.36 &{95.38} &{\bf 98.85} \\
 C6   & 95.90 &98.63     &97.49 & {\bf 100.0}&91.80 &99.09  &97.95 &98.63 &98.40 &{99.77} \\
 C7   & 75.60 &74.51     &84.20 & 79.30&87.58 &63.94  &51.85 &63.94 &72.66 &{\bf 90.52} \\
 C8   & 59.02 &62.66     &73.57 & 68.07&74.28 &68.40  &62.45 &{ 83.95} &71.63 &{\bf 84.45} \\
 C9   & 76.77 &69.68     &75.18 & 70.21&75.35 &73.40  &41.13 &73.58 &50.18 &{\bf 88.12} \\
 C10   &99.38 &99.38     &100.0 & 100.0&99.38 &98.77  &96.30 &99.38 &98.15 &{\bf 100.0} \\
 C11   &93.33 &93.25      &93.41 & 76.94&93.89 &88.83  &91.00 & {\bf 97.19} &91.94 &{94.21} \\
 C12   &73.94 &96.06      &80.61 & 90.30&63.03 &46.06  &52.12 &64.55 &86.63 &{\bf 97.58} \\
 C13   &100.0 &100.0      &100.0 & 100.0&100.0 &97.78  &95.56 &97.78 &95.45 &{\bf 100.0} \\
 C14   &87.18 &89.74      &{97.44} & 92.31&66.67 &46.15  &48.72 &76.92 &82.05 &{\bf 97.44} \\
 C15   &100.0 &90.91      &100.0 & 100.0&100.0 &72.73  &81.82 &100.0 &100.0 &{\bf 100.0} \\
 C16   &100.0 &100.0      &100.0 & 100.0&100.0 &80.00  &100.0 &100.0 &100.0 &{\bf 100.0} \\
                                                                                                             
 \hline\hline
 {OA(\%)}   & 76.53   &  77.22   &82.97  & 78.72&81.11 &74.71 &65.16 &83.38  &80.29  &{\bf 91.07} \\
 
 {AA(\%)}   & 86.02   &  86.83   &90.19  & 88.15&84.97 &77.47 &70.21 &86.49  &86.61  &{\bf 95.45} \\
 
 {$\kappa$(\%)}&73.42 & 74.21    &80.65  & 75.81&78.51 &71.21 &60.26 &80.93  &77.40  &{\bf 89.79} \\
\bottomrule[1pt]
\end{tabular}
\end{center}
\end{table*}

\begin{table*}[t]
\setlength{\aboverulesep}{-0.1pt}
\setlength{\belowrulesep}{-0.1pt}
\begin{center}
\caption{Classification accuracies (OA, AA, and $\kappa$) of different methods achieved on the Houston 2013 data.}
\label{table:houston2013_comparison}
\begin{tabular}{ c || c | c c c | c | c | c c c || c}
\toprule[1pt]
\multirow{2}{*}{Methods}     &  \multirow{2}{*}{SVM} &  \multicolumn{3}{c|}{CNNs} &  \multirow{2}{*}{RNN} & \multirow{2}{*}{miniGCN} & \multicolumn{3}{c||}{Transformers} &\multirow{2}{*}{AdverDecom} \\
 \cmidrule(lr){3-5} \cmidrule(lr){8-10}
    &   &  { 3-D CNN}&  {PResNet }&  { HybridSN} &   &  & { ViT} & {SpectralFormer} & SSFTTNet & \\
\hline\hline
 C1   &  82.62  & 83.76  & 81.67&83.57 & 81.67&{\bf 96.20}  &82.53 &83.29 & 83.29& 82.62\\
 C2   &  98.78  & 95.49  & 99.91&{\bf 100.0} & 95.39&96.90  &99.06 &98.97 &90.51 & 99.53\\
 C3   &  90.30  & 95.05  & 90.89&98.02 & 95.05&{99.41}  &91.49 &96.63 &98.61 & {\bf 99.60}\\
 C4   &  97.06  & {99.24}  & 86.74&95.55 & 96.02&97.63  &95.64 &96.02 &96.97 & {\bf 99.81}\\
 C5   &  99.81  & 99.43  & 99.43&99.72 & 97.63&97.73  &99.34 &{\bf 100.0} &99.53 & 97.82\\
 C6   &  82.52  & 90.21  & 92.31&{95.80} & 91.61&95.10  &94.41 &94.40 &91.61 & {\bf 96.50}\\
 C7   &  89.65  & 86.85  & 90.49&90.67 & 89.92&65.86  &{\bf 91.04} &83.21 &67.91 & 86.10\\
 C8   &  57.74  & {\bf 82.05}  & 75.31&81.01 & 70.09&65.15  &60.68 &80.72 &55.08 & {80.72}\\
 C9   &  61.19  & 76.49  & {80.93}&81.59 & 73.84&69.88  &71.20 &77.43 &54.25 & {\bf 85.93}\\
 C10   & 67.66   & 53.96  &70.27 &46.33 &65.93 &67.66  &52.51 &58.01 &{81.56} & {\bf 83.49}\\
 C11   & 72.68	   & 82.35  &84.91 &{\bf 94.12} &70.40 &82.83  &78.75 &80.27 &90.51 & 89.66\\
 C12   & 70.41   & 78.48  &71.85 &80.50 &79.73 & 68.40 &81.27 &84.44 &{\bf 84.73} & {81.94}\\
 C13   & 61.05   & 75.44  &89.47 &{\bf 94.74} &74.39 & 57.54 &65.96 &73.33 &81.75 & 92.98\\
 C14   & 94.33   & 91.90  &97.57 &96.36 &98.79 & 99.19 &95.14 &{\bf 99.60} &99.19 & 99.19\\
 C15   & 80.13   & 92.18  &{\bf 100.0} &95.78 &98.31 & 98.73 &92.39 &99.15 &99.79 & {98.94}\\
                                                                                                             
 \hline\hline
 {OA(\%)}      & 80.16  & 84.71 & 85.59 &86.89 &83.55 &82.31 &82.22 & 85.55 & 82.46 & {\bf 90.03}\\
 
 {AA(\%)}      & 80.40  & 85.53 & 87.45 &{88.92} &85.25 &83.88 &83.43 & 87.03 & 85.02 & {\bf 91.66}\\
  
 {$\kappa$(\%)} & 78.44   & 83.40 & 84.35 &85.77 &82.15 &80.84 &80.68 & 84.32 & 80.97 & {\bf 89.18}\\
\bottomrule[1pt]
\end{tabular}
\end{center}
\end{table*}

\section{Conclusions}\label{sec:conclusions}

In this work, based on intrinsic property of hyperspectral image, we develop a deep intrinsic decomposition with adversarial learning for hyperspectral image classification. We develop the adversarial network to decompose the learned feature into the category-related and environmental-related features. Then, based on the proposed adversarial learning methods, the network can be adversarially learned and provide discriminative features of the hyperspectral image. Experimental results over different CNN backbone shows that the proposed method can remarkably improve the classification performance. Besides, the comparison results between other state-of-the-art methods also show the superiority of the proposed method.

In future works, it would be interesting to investigate the effects of the proposed AdverDecom on the applications of other tasks, such as anomaly detection, target identification. Besides, exploring the performance of AdverDecom by integrating other training strategies, such as metric learning, is another interesting future topic.

\ifCLASSOPTIONcaptionsoff
  \newpage
\fi



%

\bibliographystyle{IEEEtran}
\bibliography{references}




%




%


\vfill


\end{document}